\documentclass[conference]{IEEEtran}
\usepackage{algorithm}      
\usepackage{algpseudocode} 
\usepackage{caption} 
\usepackage[absolute]{textpos} 
\usepackage{marvosym}
\usepackage{multirow}
\usepackage{amsmath}
\usepackage{makecell}
\usepackage{tikz}
\usepackage{booktabs} 
\usepackage{rotating}
\usepackage{xspace}
\newcommand{\eg}{\textit{e.g.,}\@\xspace}
\newcommand{\ie}{\textit{i.e.,}\@\xspace}
\usepackage{xcolor}
\usepackage{url}
\usepackage{cite}
\usepackage[numbers,sort&compress]{natbib}
\IEEEoverridecommandlockouts
\usepackage{subcaption}
\usepackage[table]{xcolor}
\usepackage{makecell}
\def\BibTeX{{\rm B\kern-.05em{\sc i\kern-.025em b}\kern-.08em
    T\kern-.1667em\lower.7ex\hbox{E}\kern-.125emX}}
\begin{document}
\pdfpagewidth=8.5in
\pdfpageheight=11in

\title{
SMoE: An Algorithm-System Co-Design for Pushing MoE to the Edge via Expert Substitution
}

\author{
    \IEEEauthorblockN{
        Guoying Zhu\IEEEauthorrefmark{1}, 
        Meng Li\IEEEauthorrefmark{1}\textsuperscript{\Letter}, 
        Haipeng Dai\IEEEauthorrefmark{1}\textsuperscript{\Letter}, 
        Xuechen Liu\IEEEauthorrefmark{1}, 
        Weijun Wang\IEEEauthorrefmark{2}, 
        Keran Li\IEEEauthorrefmark{1}, \\
        Jun Xiao\IEEEauthorrefmark{3}, 
        Ligeng Chen\IEEEauthorrefmark{3}, and 
        Wei Wang\IEEEauthorrefmark{1}
    }
    \IEEEauthorblockA{
        \IEEEauthorrefmark{1}State Key Laboratory for Novel Software Technology, Nanjing University,
        \IEEEauthorrefmark{2}Tsinghua University,,
        \IEEEauthorrefmark{3}Honor Device Co., Ltd
    }
    \IEEEauthorblockA{
        Email: 522023330124@smail.nju.edu.cn, \{meng, haipengdai\}@nju.edu.cn, 
        522025330063@smail.nju.edu.cn, \\
        wangweijun@air.tsinghua.edu.cn, keranli@smail.nju.edu.cn, 
        sunny-xiaojun@hotmail.com, \\
        \{chenlg@smail, ww\}@nju.edu.cn
    }
    \thanks{\textsuperscript{\Letter}Corresponding authors.}
}

\maketitle
\begin{abstract}
The Mixture of Experts (MoE) architecture has emerged as a key technique for scaling Large Language Models by activating only a subset of experts per query. 
Deploying MoE on consumer-grade edge hardware, however, is constrained by limited device memory, making dynamic expert offloading essential. 
Unlike prior work that treats offloading purely as a scheduling problem, we leverage expert importance to guide decisions, substituting low-importance active experts with functionally similar ones already cached in GPU memory, thereby preserving accuracy. 
As a result, this design reduces memory usage and data transfer, while largely eliminating PCIe overhead.
In addition, we introduce a scheduling policy that maximizes the reuse ratio of GPU-cached experts, further boosting efficiency.
Our extensive evaluations show that, compared with state-of-the-art approaches, our method achieves a 48\% reduction in decoding latency and maintains an expert cache hit rate above 60\%, all while preserving nearly lossless accuracy.
\end{abstract}


\section{Introduction}
\textbf{MoE architectures offer a promising approach for deploying Large Language Models (LLMs) on edge devices, addressing an increasingly critical need \cite{shazeer2017outrageously, sarkar2023edge, kang2025flame}}.
Edge applications such as smart homes \cite{rivkin2024aiot}, intelligent healthcare \cite{awasthi2025artificial}, autonomous transportation \cite{zhang2024advancing}, and pervasive video analytics \cite{wang2025region} demand low latency and strong data privacy, which makes edge deployment essential.
These applications run in a low-batch regime, where latency is more critical than throughput, since an edge LLM usually serves only one device instead of a cluster with high concurrency \cite{li2025h2,friha2024llm,spector2023accelerating}.
Yet, edge servers are often limited in computational capacity and GPU memory, restricting full model deployment and rapid inference \cite{singh2023edge, yu2024cambricon}. 
Compared with dense models that compute all parameters for every input, MoE architectures mitigate these constraints by partitioning feed-forward layers into multiple experts~\cite{fedus2022switch}, activating only a sparse subset per token. 
This design drastically reduces computational overhead.

\begin{figure}[t]
    \begin{minipage}{\linewidth}
        \centering
        \includegraphics[width=\linewidth]{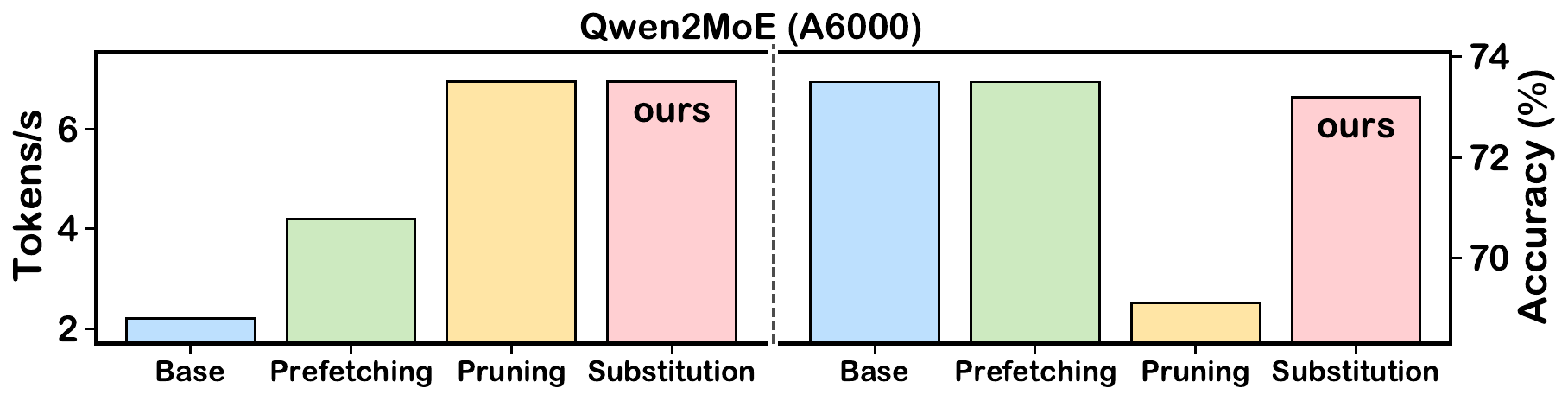}
        \caption{Token generation speed and accuracy vary among prefetching, expert pruning and our substituting methods.} \label{introshow}
    \end{minipage}
\end{figure}

\textbf{However, GPU memory limits force frequent offloading of experts to CPU memory and reloading to the GPU, causing significant inference latency.} 
Since edge GPUs have limited memory and cannot hold all experts simultaneously, inference often requires offloading experts to slower CPU memory or leveraging the CPU to perform computations \cite{jiang2024neo,zhang2025daop,kim2025lia}. 
Both PCIe transfers and CPU computation are significantly slower than GPU execution, introducing $10–100\times$ higher latency.
In particular, to mitigate prolonged CPU-GPU expert loading latency, existing solutions fall into two categories: (1) expert prefetching and (2) expert pruning.
Prefetching-based works, including MoE-infinity \cite{xue2024moe}, HybriMoE \cite{zhong2025hybrimoe}, and ProMoE~\cite{song2024promoe}, use prefetching instructions to overlap expert loading and computation, hiding CPU-GPU transfer latency.
The reduced latency depends on the degree of overlap and is often limited.
In contrast, pruning-based approaches~\cite{pruning,lu2024not} reduce latency by directly dropping some experts, which may degrade LLM accuracy due to incomplete expert computation.
Moreover, as shown in Fig.~\ref{introshow}, to evaluate the effectiveness of these two approaches, we conduct a micro-benchmark on a human examination dataset, including subjects from Math, History, and Biology in the Gaokao benchmark~\cite{allenai:arc}.
It can be easily observed that prefetching-based works suffer from high token generation latency, while pruning-based approaches suffer from reduced accuracy.
This raises a fundamental yet unanswered question: how can we design a GPU-friendly expert scheduling mechanism that reduces inference latency without compromising model accuracy?

\textbf{To address this challenge, we propose a novel, third category of methods: expert substitution.}
This method is motivated by an insightful observation: in fine-grained MoE models like Qwen \cite{qwenmoe} or DeepSeek \cite{deepseekmoe}, although top-$k$ experts may be activated at a given time, only a small subset achieves high scores (termed \emph{top-score active experts}), while the rest receive scores comparable to inactive experts (\emph{low-score active experts}).
For these low-score experts, substituting them with other experts having similar scores preserves accuracy or incurs only limited degradation. 
This model-level insight opens a system-level optimization opportunity for expert offloading: a low-score activated expert can be replaced with a similar-scoring expert already cached on the GPU.
Moreover, this substitution-based method is fully compatible with prefetching-based techniques and can even strengthen them, as it simultaneously reduces the number of experts that need to be prefetched.
As Fig.~\ref{introshow} shows, the substitution-based method achieves comparable accuracy with faster token generation than pure prefetching, and comparable generation speed with higher accuracy than pruning on the Gaokao benchmark~\cite{allenai:arc}. 

\begin{figure}[t]
    \begin{minipage}{\linewidth}
        \centering
        \includegraphics[width=\linewidth]{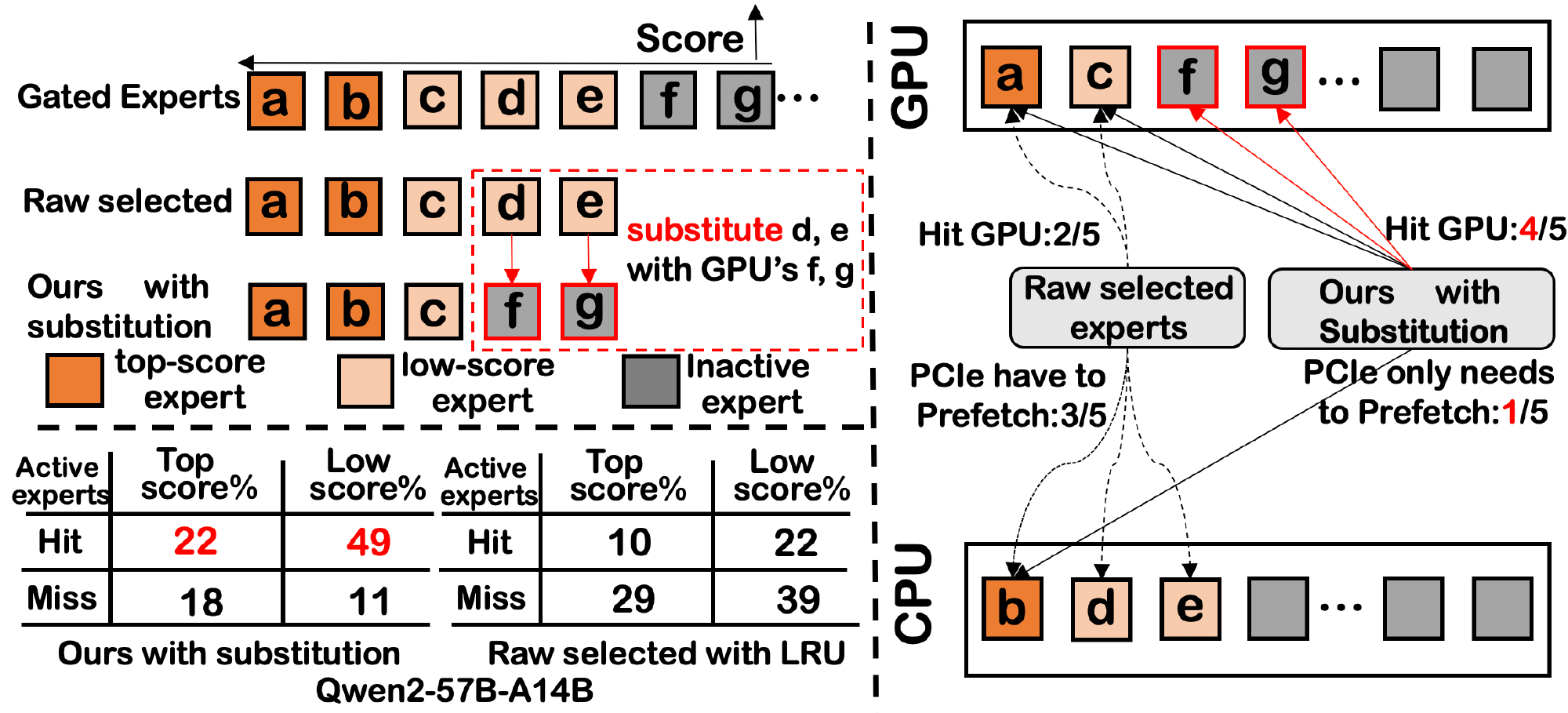}
        \caption{
        Naive MoE layer-wise expert loading vs. our substitution-centric expert scheduler.
        %
        } \label{intro}
    \end{minipage}
\end{figure}
%
%
%
To better illustrate our general idea, we provide a simple running example in Fig.~\ref{intro}, which further demonstrates that this strategy largely eliminates the CPU computation burden of low-score active experts while ensuring that top-score active experts are loaded from CPU memory to the GPU in advance.
Particularly, by substituting low-score experts $d$ and $e$ with GPU-resident experts $f$ and $g$ of similar scores, the GPU expert hit rate per layer increases from $\frac{2}{5}$ to $\frac{4}{5}$.
Besides, this substitution method reduces prefetching overhead from three experts to one and further enhances the GPU expert hit rate to 71\% for the Qwen2-57B-A14B model on an A6000~\cite{qwenmoe}.

\vspace{1mm}
\noindent
\textbf{Challenges.} 
Our work addresses three main challenges: 
\begin{itemize}
\setlength{\leftskip}{-0.4cm}
\item How to identify low-score experts and suitable substitutes in GPU memory with minimized accuracy loss?
\item How to decide which experts to evict or prefetch, given that cached experts can serve as substitutions, to maximize the overall cache hit rate?
\item How to integrate CPU-assisted pipelines to handle experts that can be neither substituted nor prefetched?
\end{itemize}

\vspace{1mm}
\noindent
\textbf{Contributions.} 
Our contributions are as follows.
\begin{itemize}
\setlength{\leftskip}{-0.4cm}
\item 
We introduce the SMoE (Expert Scheduler with Substitution) to minimize decoding latency by maximally substituting low-importance active experts with functionally similar ones already cached in GPU memory, while incurring almost no loss in accuracy.
\item 
We design a CPU-GPU-load pipeline system specifically for MoE LLMs with SMoE, capable of handling online workloads without requiring any offline preparation.
\item 
We extensively evaluate our approach on practical workloads, demonstrating its effectiveness and efficiency.
\end{itemize}

\section{Background}
Deploying Large Language Models (LLMs) directly on edge devices is crucial for instant, private, secure, and reliable interaction within user environments like smart homes and autonomous vehicles, demanding real-time processing and immediate perception \cite{friha2024llm,li2025moe,kong2025serving}.
For LLM on edge, Time Per Output Token (TPOT) \cite{nvidiabentch} is a critical metric that measures the average time between successively generated tokens during decoding.
Edge inference typically handles low-batch requests \cite{friha2024llm}, since an edge LLM serves an individual device rather than a computing cluster. In such low-batch scenarios, performance is often constrained by memory bandwidth as in expert offloading, making TPOT optimization essential.

To address the limited GPU memory on edge hardware, MoE LLMs with online expert offloading offer a solution by segmenting traditional layers into specialized experts \cite{fedus2022switch}. 
This MoE architecture processes only a sparse subset of parameters per input, significantly easing the strain on constrained GPU resources compared to dense networks.

Our goal is to reduce the decoding latency of LLMs with MoE that feature fine-grained expert segmentation on GPU-memory constrained edge devices. 
We discuss our background from three perspectives:
(1) the model architecture (MoE with fine-grained expert segmentation),
(2) the online expert offloading architecture, and (3) the metric TPOT (Time Per Output Token), measuring the latency time between successively generated tokens during decoding.

\subsection{LLM with Fine-grained Expert MoE}
%

The MoE LLM substitutes the dense FFN with multiple smaller expert networks \cite{cai2025survey}. For each input, a router selects the top‑k experts to activate, significantly improving parameter efficiency by computing only through these chosen modules.

Unlike traditional MoE that activates 1–2 experts per token, fine-grained MoE divides experts further and activates more per token under similar computational constraints. This promotes specialization and allows richer integration of expert knowledge, enhancing model expressiveness.
DeepSeekMoE first introduced this strategy \cite{dai2024deepseekmoe}, later adopted by models like Qwen2-57B-A14B \cite{qwenmoe} and XVERSE-MoE-A4.2B \cite{xversemoe}, achieving strong performance with reduced training costs.
%

MoE with fine-grained expert segmentation often includes shared experts that process all tokens \cite{deshpande2024moesaic}, regardless of routing decisions.
This design tackles a major issue in traditional MoE, where common knowledge is redundantly stored. 
By using shared experts to consolidate common knowledge, the architecture allows other experts to specialize in distinct domains. This division enhances parameter efficiency.

\begin{figure}[t]
    \begin{minipage}{\linewidth}
        \centering
        \includegraphics[width=0.9\linewidth]{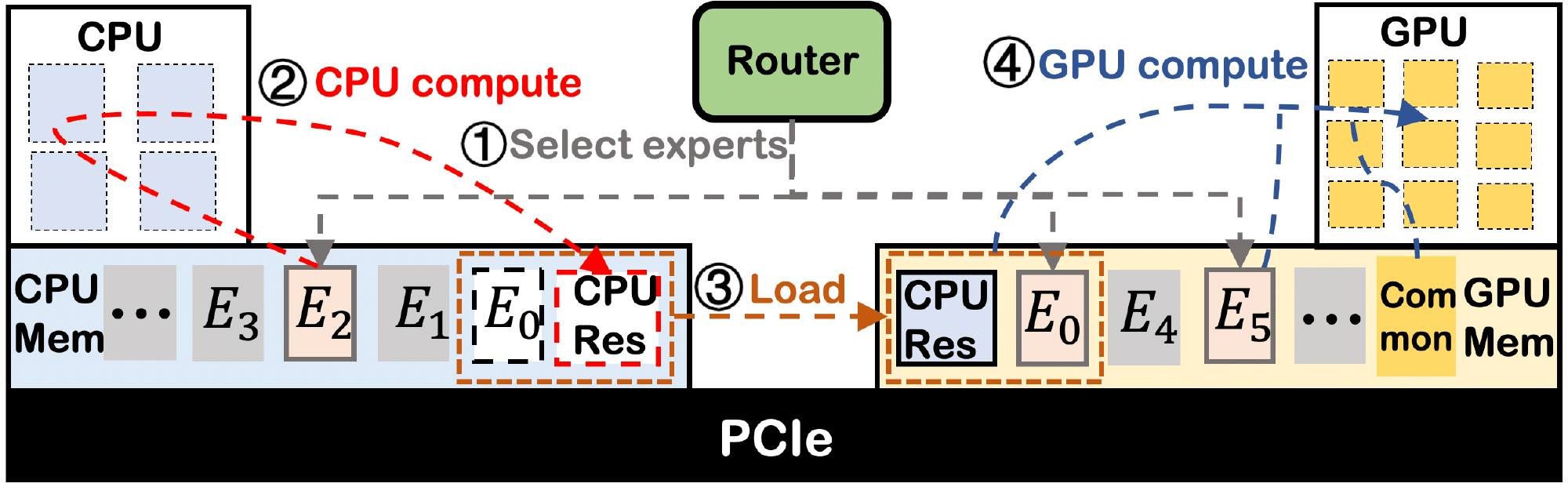}
        \caption{Online Expert Offloading in MoE LLMs at one layer. Step 1: Router selects the active experts. Step 2: CPU computes part of the active experts in CPU memory. Step 3: Part of active experts and CPU-computed expert results are transferred to GPU memory via PCIe. Step 4: GPU processes experts, consolidating those results with CPU-computed results.} \label{expertoffload}
    \end{minipage}
\end{figure}
\begin{figure*}[t]
    \begin{minipage}{0.49\linewidth}
        \centering
        \includegraphics[width=\linewidth]{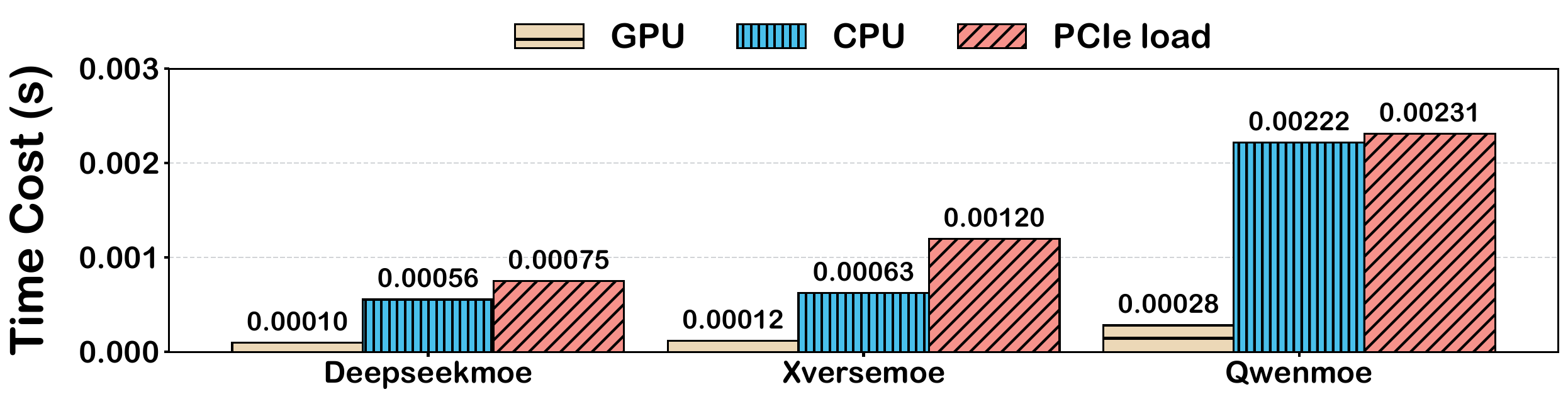}
        \caption{CPU/GPU (A6000) time for expert-token computing, \& PCIe time for expert loading from 3 MoE LLMs.} \label{GPUCPUload}
    \end{minipage}
    \hfill
    \begin{minipage}{0.49\linewidth}
        \centering
        \includegraphics[width=\linewidth]{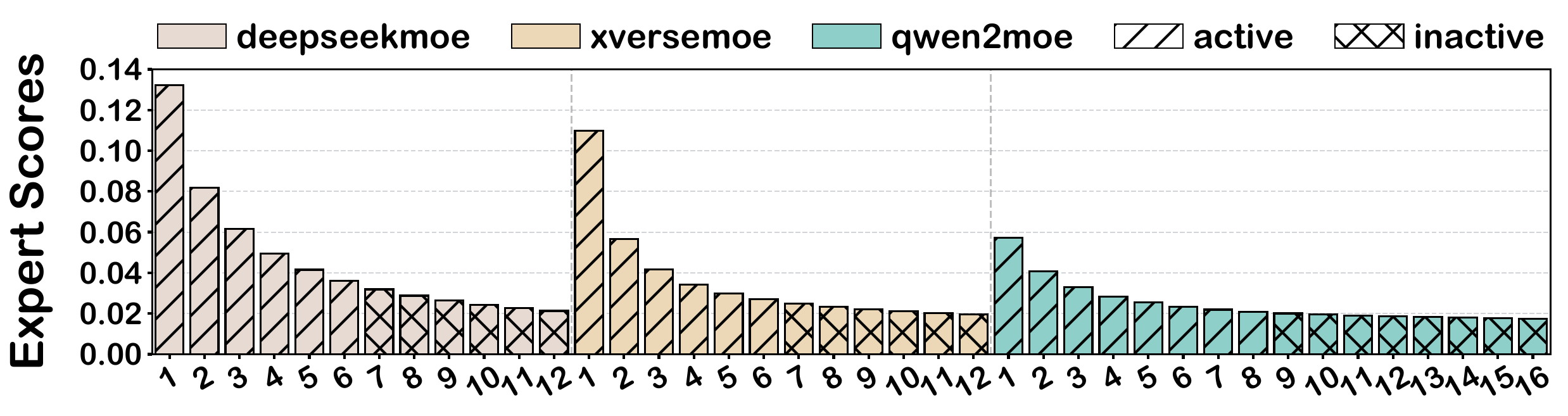}
        \caption{Average scores at each ranked position across all tokens and layers (sorted in descending order) from 100 prompts.}  \label{obser}
    \end{minipage}
\end{figure*}

\subsection{Expert Offloading in MoE LLMs}

Expert Offloading is crucial for deploying MoE LLMs on edge devices, operating at the expert level rather than the layer level, unlike LLM offloading \cite{song2024powerinfer, zhang2024edgeshard}. 
It strategically allocates a subset of experts and common parameters, such as attention, token embedding, and router weights to GPU memory, while storing all expert parameters in CPU memory. 
This approach substantially lowers TPOT compared to general LLM methods like llama.cpp \cite{llamacpp}, which are not tailored for MoE models.
Moreover, other techniques such as PowerInfer \cite{song2024powerinfer, xue2024powerinfer2}, which target LLMs with ReLU activation functions, do not provide specific offloading strategies for MoE models.

Similarly to the general LLM offloading technique: two main strategies are used, as shown in Steps 2 and 3 of Fig. \ref{expertoffload}: either freeing up GPU memory to transfer the necessary expert parameters from CPU memory for GPU computation, or directly performing the computations on the CPU and then aggregating the results with those from the GPU.
These two approaches can be pipelined.
As shown in Fig. \ref{expertoffload}, while the CPU computes one expert ($E_2$), another ($E_0$) can be simultaneously loaded into GPU memory via PCIe. 
Upon $E_2$'s completion, its results are then transferred to GPU memory, allowing for concurrent processing and data movement.

Our work addresses unknown edge workloads through online expert offloading.
Expert offloading is categorized into online and offline MoE serving strategies, depending on whether experts are dynamically loaded into GPU memory based on the current requests or the entire workload.
Online strategies manage dynamically changing edge requests, adapting through flexible scheduling to handle the impact of frequent loading and computation of non-resident experts on latency. 
Examples include MoE-Infinity \cite{xue2024moe} and HybriMoE \cite{zhong2025hybrimoe}. Offline strategies focus on predetermined workloads, optimizing GPU use by capturing expert activation patterns and employing expert pruning, as seen in MoE-lightning \cite{moelight}.
  
%




\section{Motivation}

\begin{table}[b]
\centering
\setlength{\tabcolsep}{3pt}
\caption{TPOT time ratio distribution for Qwen on A6000.}
\label{ratio}
\begin{tabular}{ccc}
\toprule
Low-score loading & Top-score loading & GPU computing \\
\midrule
42\% & 29\% & 29\% \\
\bottomrule
\end{tabular}
\label{tab:time_ratio}
\end{table}
Deploying MoE LLMs on edge devices necessitates online expert offloading due to constrained GPU memory.
As workloads dynamically shift, so does the set of active experts; however, the limited on-device GPU memory cannot always accommodate all required experts.
Consequently, active experts residing in GPU memory are processed directly, while those in CPU memory are either transferred to the GPU for computation or processed directly on the CPU. 
However, processing experts not already in GPU memory significantly impacts inference latency, regardless of the offloading method.
As Fig. \ref{GPUCPUload} shows, this is primarily because PCIe loading can be one to two orders of magnitude slower than GPU computation, and CPU computation is typically one order of magnitude slower.
Thus, deployment of MoE LLMs under GPU memory constraints requires minimizing overhead from expert loading.

Several optimization strategies have emerged for online expert offloading to alleviate bottlenecks. 
One approach increases the hit rate of experts in GPU memory through prefetching and caching. Prefetching predicts future needs, enabling preemptive loading to reduce TPOT impact \cite{xue2024moe}, while caching strategies like LRU minimize data transfers. Another line of work adopts a related idea by optimizing CPU computation through efficient CPU-side loading pipelines \cite{zhong2025hybrimoe}.

\begin{figure}[b]

        \begin{minipage}{\linewidth}
        \centering
        \includegraphics[width=0.9\linewidth]{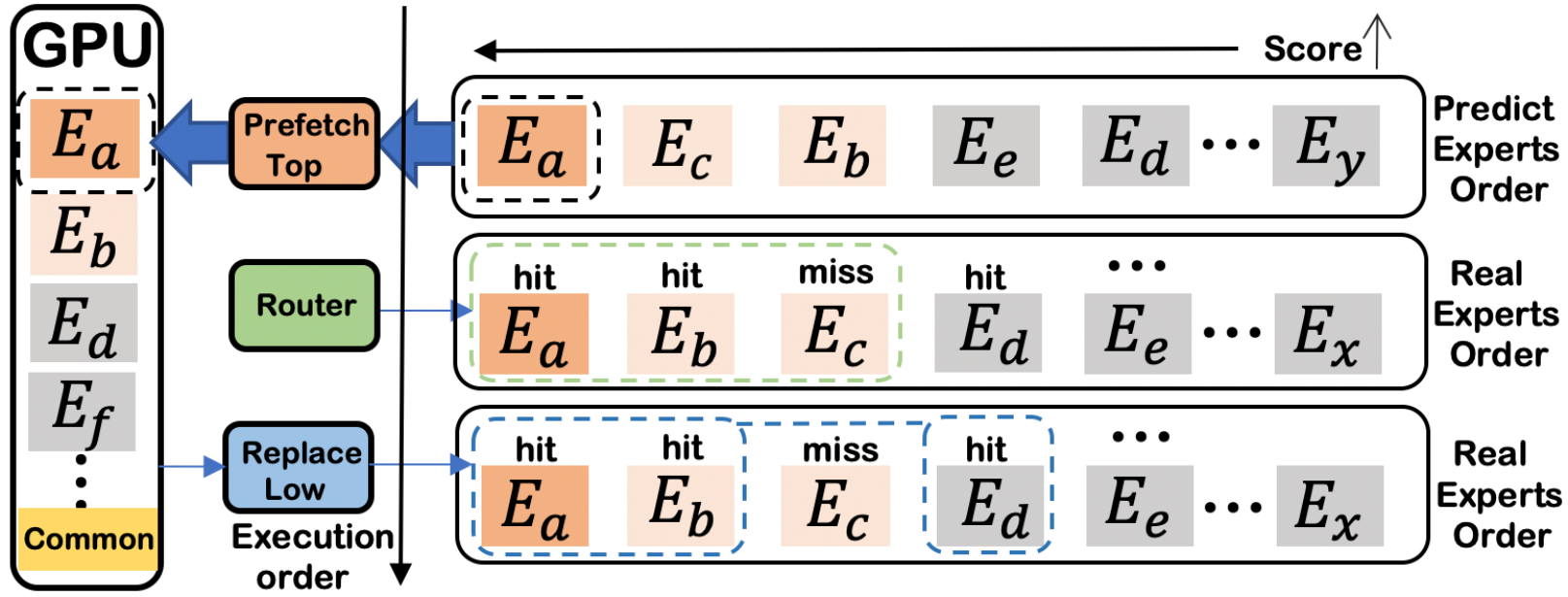}
        \caption{Our idea: prefetching top-score experts and replacing low-score experts in each iteration at one layer.} \label{idea}
    \end{minipage}
\end{figure}
However, current offloading strategies overlook the significant variation in importance among activated experts, scheduling them uniformly.
In MoE architectures, an expert's importance to an input is reflected in its router's gate score, with higher scores indicating greater significance.
As Fig. \ref{obser} illustrates, a distinct pattern of importance scores emerges among activated non-shared experts: only a few achieve high scores (top-score experts), significantly influencing the output, while others have low scores (low-score experts), similar to inactive experts. 
This differentiation arises because (1) shared experts handle common knowledge and (2) fine-grained segmentation creates highly specialized non-shared experts.
Yet, existing online expert offloading methods, such as prefetching and CPU-load pipelining \cite{zhong2025hybrimoe}, do not consider this pattern of activated experts on output results.
This oversight results in time-consuming operations, like CPU computation and PCIe loading of experts, being used for experts that have minimal impact on the final outcome. However, Table \ref{ratio} shows that the low-scoring loading time is the main bottleneck of TPOT.

Our proposed online strategy schedules experts by their importance, substituting low-importance active experts (low-score experts) with functionally similar ones already cached in GPU memory. This mechanism relies on the observation that routing redundancy and noisy gating in sparse MoEs cause tail experts with marginal scores to behave similarly. Previous research indicates that MoE routing exhibits inherent stochasticity because training procedures inject routing noise and enforce load balancing to distribute tokens across the expert pool \cite{shazeer2017outrageously, fedus2022switch}. Consequently, tail experts within the selected top-k set usually receive marginal scores and demonstrate unstable selection frequencies. These experts generally provide weak or redundant signals. Swapping them with cached alternatives rarely leads to noticeable accuracy drops and may actually prevent the dilution of signals from highly confident experts.

As Fig. \ref{idea} illustrates, we first prefetch top-score experts based on predicted expert scores, leveraging pipelining to overlap their loading time with computation.
%
%
Since PCIe loading time significantly exceeds computation time, as shown in Fig. 4, this prefetching can only ensure timely resource access for critical top-score experts.
After the gating network determines the actual expert scores and the top-score experts (\eg $k=3$) are selected, a top-score expert like $E_a$ may already be prefetched while a remaining expert like $E_c$ resides in CPU memory.
This typically requires a PCIe transfer or CPU computation. Unlike previous methods, our approach recognizes the low score and minimal output impact of $E_c$. We therefore substitute $E_c$ with a similarly scored and GPU-resident inactive expert such as $E_d$.
During this process, our scheduler strictly prioritizes the highest-scoring cached candidate whose score falls within a defined threshold.
The system specifically attempts to find the closest ranked inactive expert first. 
If no cached expert satisfies this rigorous score boundary, the scheduler skips the substitution and falls back to either loading the originally selected expert via PCIe or computing it directly on the CPU.

Moreover, although identifying the optimal substitute requires scanning the residing experts in the same layer, this operation incurs no impactful system delay due to the extremely small search space per layer.
This lightweight search is executed on the CPU and fully pipelined with ongoing GPU computations and PCIe load operations.
This allows almost all the selected experts to be computed directly on the GPU, effectively mitigating the TPOT impact of low-score experts at a negligible cost to accuracy.

\section{Expert Scheduler with Substitution}
This section outlines the design objectives and approach of our scheduler with expert substitution (SMoE), which aims to achieve three key functions.
First, it performs substitution of low-score experts, identifying and replacing them to maximize the GPU expert cache efficiency. 
Second, it implements top-score expert prefetching to overlap loading and computation times, mitigating accuracy loss from replacement.
Third, it uses CPU-assisted computing to dynamically decide whether to transfer active experts to the GPU via PCIe or compute them directly on the CPU, addressing prefetching failures and expert-cache router limitations.
\begin{table}[t]
\centering
\caption{Definitions.}\label{definition}
\begin{tabular}{cl}
\toprule
\textbf{Symbol} & \textbf{Description} \\
\midrule
\( G \) & Experts already in the GPU. \\
\( E_a \) & Experts selected by top-k. \\
\( E_l \) & Low-score experts. \\
\( E_{a} \setminus E_l \) & Top-score experts. \\
\( E_s \) & Experts in \( G \) that can serve as substitutes. \\
\( E_p \) & Experts prefetched from the previous layer. \\
\(\alpha \) & Expert substitution threshold.\\
$S_{k+1}$ & The gate score of the $k+1$-th highest expert.\\
\bottomrule
\end{tabular}
\label{tab:expert_sets}
\end{table}
\subsection{Design Analysis}\label{design}
As discussed earlier, the absence of experts in GPU memory significantly increases inference latency, which can be mitigated by increasing the hit rate of experts cached in the GPU.
Our system improves this hit rate by performing expert substitution independently for each layer, rather than considering all experts at once. 
This design choice reflects the impracticality of jointly analyzing all layers in transformer LLMs, because the experts chosen in one layer can strongly influence subsequent layers.

Our goal is to maximize, for each layer, the number of routing hits on experts that already reside in GPU memory. To analyze this effect, we define several expert sets and measure their hit counts at the layer level, as shown in Table \ref{definition}.
Our optimization objective, given by Equation \eqref{op}, consists of the number of experts already in the GPU and the number of low-score experts replaced by experts in the GPU.
\begin{equation}\label{op}
    \max |G \cap E_a| + min(|E_l \setminus G|, |E_s|) 
\end{equation}
It can be expanded as \( |(G \setminus E_p \cup E_p) \cap E_a| + \min(|E_l \setminus (G \setminus E_p \cup E_p)|, |E_s|) \), or decomposed into Equation \eqref{dec}.
\begin{equation}\label{dec}
\small
\max \ \ |(G \setminus E_p) \cap E_a| + |E_p \cap E_a| + min(|E_l \setminus (G \setminus E_p)|, |E_l \setminus E_p|, |E_s|)
\end{equation}
Here, \( G \setminus E_p \) and \( E_a \) are constant, as they depend on pre-existing expert distributions in the GPU and the top-k selection results by the gate.
\textbf{\textit{C$_1$, C$_2$}}: Thus, expanding \( E_l \) and \( E_s \) becomes a viable and beneficial optimization direction. A broader \( E_s \) provides more potential candidates for substitution, while a larger \( E_l \) increases the pool of GPU-resident substitutes, both directly contributing to raising the hit-related terms in our optimization objective. Regarding \( E_p \), increasing it can reduce \( |E_l \setminus E_p| \) but can increase \( |E_p \cap E_a| \).%
\textbf{\textit{C$_5$, C$_6$}}: We increase \( E_p \) because the growth in \( |E_p \cap E_a| \) directly enhances the overall optimization target, whereas the change in \( |E_l \setminus E_p| \) does not necessarily contribute positively to the minimization.
While increasing \( E_p \) and \( E_l \), we aim to maximize \( |E_l \setminus E_p| \), which means minimizing \( |E_p \cap E_a| \). \textbf{\textit{C$_4$}}:Therefore, we prioritize prefetching the top-score experts.

However, our system needs to apply constraints to \( E_l \) and \( E_s \) to balance MoE model with GPU hit rate improvement. The scores by the gate represent an expert's importance to the output, so aligning the scores of the substituted experts with those of the low-score experts can prevent a significant drop in accuracy.
We introduce a hyperparameter \(\alpha\) (expert substitution threshold) that acts as the score threshold to determine which experts in \( E_s \) and \( E_l \) are eligible for replacement.
Specifically, \( S_{k+1} \) is the gate score of the \( k+1 \)-th highest expert. The constraints are: 
$$
S_{k+1} < \text{Score}(e) < (1+\alpha) S_{k+1}, \quad e \in E_l,
$$
$$
(1-\alpha) S_{k+1} < \text{Score}(e) \le S_{k+1}, \quad e \in E_s.
$$

These bounds ensure the scores of \( E_l \) and \( E_s \) are closely aligned with each other. We aim for \( E_l \) and \( E_s \) to meet these constraints.
The constraint on \( E_p \) depends on the time available for loading experts from the previous layer's prefetching start to the current layer's expert computation. 
\textbf{\textit{C$_5$}}:Thus it requires that the cost of prefetching be sufficiently low to save time and allow for the loading of more experts.

Based on the design analysis, our system's implementation criteria are as follows:
\begin{itemize}
\setlength{\leftskip}{-0.4cm}
\item 
(1) \textbf{\textit{C$_1$}: Maximizing $E_l$. }
Low-score experts should constitute a significant proportion of the active experts.
\item 
(2) \textbf{\textit{C$_2$}: Maximizing $E_s$. }The GPU memory must contain an increased inactive experts suitable for substitution.
\item 
(3) \textbf{\textit{C$_3$}: Constraints on $E_l$ and $E_s$. }The scores of low-score experts must be comparable to those of certain inactive experts within the GPU to enable smooth substitution. 
\item 
(4) \textbf{\textit{C$_4$}: Minimizing \( |E_l \cap E_p| \). }Prioritize prefetching top-score experts.
\item 
(5) \textbf{\textit{C$_5$}: Maximizing $E_p$. }Begin prefetching experts for the next layer as quickly as possible to maximize the number of prefetched experts.
\item 
(6) \textbf{\textit{C$_6$}: Maximizing $E_p$. }Ensure the prefetching accuracy.
\end{itemize}

\subsection{Low-score Expert Substitution}

\begin{algorithm}
\caption{Expert-Cache Router}
\label{expertcacherouter}
\begin{algorithmic}
\Require $\alpha$, $k$, $S$ \Comment{Score inputs}
\Ensure $O$ \Comment{Output cacherouter\_experts}
\State Initialize $O$, set thresholds $T$, $L$, $R$, sets $A$, $B$, $C$ \Comment{$A=E_s$, $B=E_l$, $C=E_a\setminus E_l$}
\For{each token $t$}
    \State Sort $S_t$ desc; $S_{k+1} \gets \text{(k+1)-th score of } S_t$
    \State $T \gets (1+\alpha)S_{k+1}$; $L \gets S_{k+1}$; $R \gets (1-\alpha)S_{k+1}$
    \For{each expert $e$}
        \If{score of $e > T$} Add $e$ to $O[t]$ and $C$
        \EndIf
    \EndFor
    \State Initialize $B_t$, $A_t$
    \For{each expert $e$}
        \If{$L \leq$ score of $e < T$} Add $e$ to $B_t$
        \ElsIf{$R \leq$ score of $e < L$ and $e$ in GPU or $C$} Add $e$ to $A_t$
        \EndIf
    \EndFor
    \If{$|A_t| \geq |B_t|$}
        \State Select top $|B_t|$ experts from $A_t$; Add to $O[t]$
    \Else
        \State Add $A_t$ to $E[t]$; Add top $|B_t| - |A_t|$ experts from $B_t$ to $O[t]$
    \EndIf
\EndFor
\State \Return $E$
\end{algorithmic}
\end{algorithm}

To fulfill these system design requirements, two key \textit{\textbf{questions}} must be addressed: first, how to categorize the active experts selected by the original MoE router, distinguishing between low-score and top-score experts to meet \textbf{\textit{C$_1$, C$_3$}}; and second, how to design a GPU cache strategy that retains a greater number of experts eligible for substitution within the GPU, aiming to meet \textbf{\textit{C$_2$}}.
We design an expert-cache router and a cache eviction to answer these two questions.

\noindent
\textbf{Expert-cache router.} Algorithm \ref{expertcacherouter} illustrates our approach for the expert-cache router. 
Those experts with scores above $(1+\alpha)S_{k+1}$ are top-score experts, while scores between $S_{k+1}$ and $(1+\alpha)S_{k+1}$ are low-score.
For top-score experts, when we process multiple requests concurrently, multiple tokens are decoded, each possessing its own set of top-score experts at the current layer.
Given their critical role in computation, these experts are retained in the results of the expert-cache router and called as the top-score expert set.
For low-score experts, if they are already in GPU memory or part of the top-score expert set, they are retained in the expert-cache router's results without incurring additional CPU overhead. 
Otherwise, they can be replaced with an inactive expert whose score falls between $(1-\alpha)S_{k+1}$ and $S_{k+1}$ and is present in the GPU or in the top-score expert set. 
We select the $|E_l\setminus G|$ highest-scoring alternatives from $E_s$ to be included in the expert-cache router's results.
If $|E_l\setminus G|$ is larger than $|E_s|$, then only $|E_l\setminus G| - |E_s|$ low-scoring experts are prepared for loading into GPU memory or for direct computation by the CPU.

\begin{figure*}
    \begin{minipage}{0.49\linewidth}
        \centering
        \includegraphics[width=0.85\linewidth]{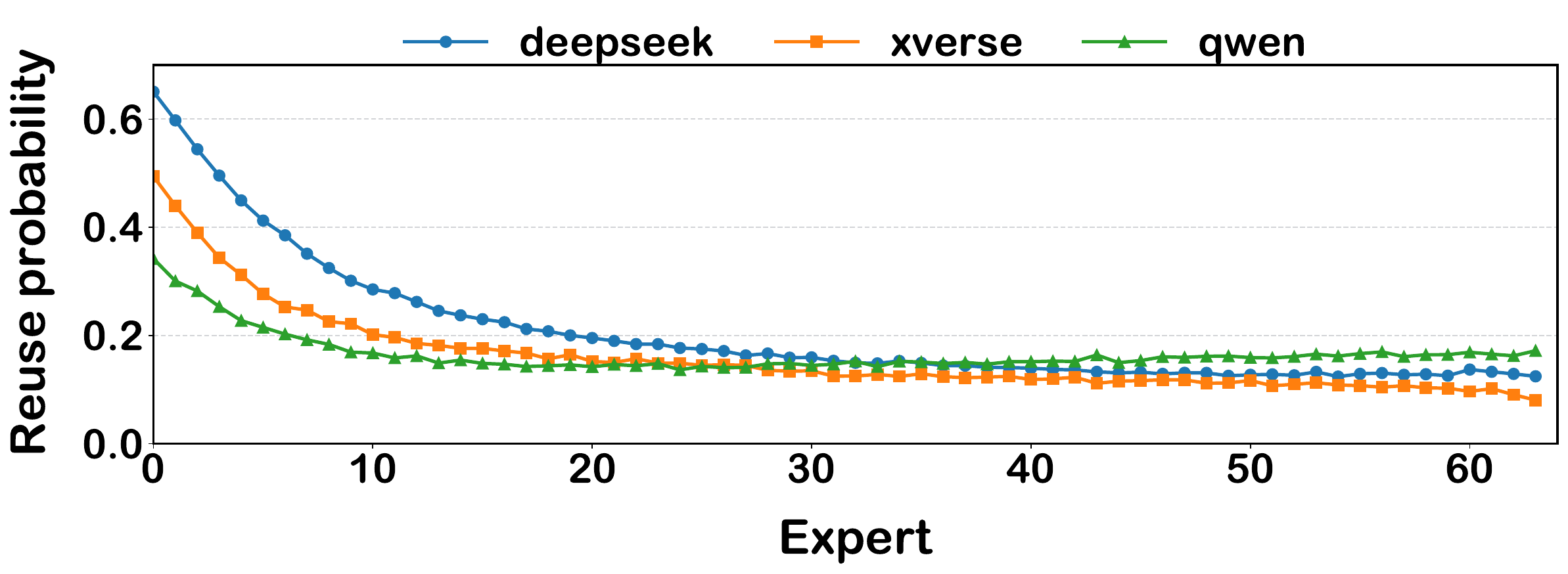}

        \caption{The reuse probability of experts based on score (in descending order) in three MoE models.} \label{expertreuse}
    \end{minipage}
    \hfill
        \begin{minipage}{0.49\linewidth}
        \centering
        \includegraphics[width=0.95\linewidth]{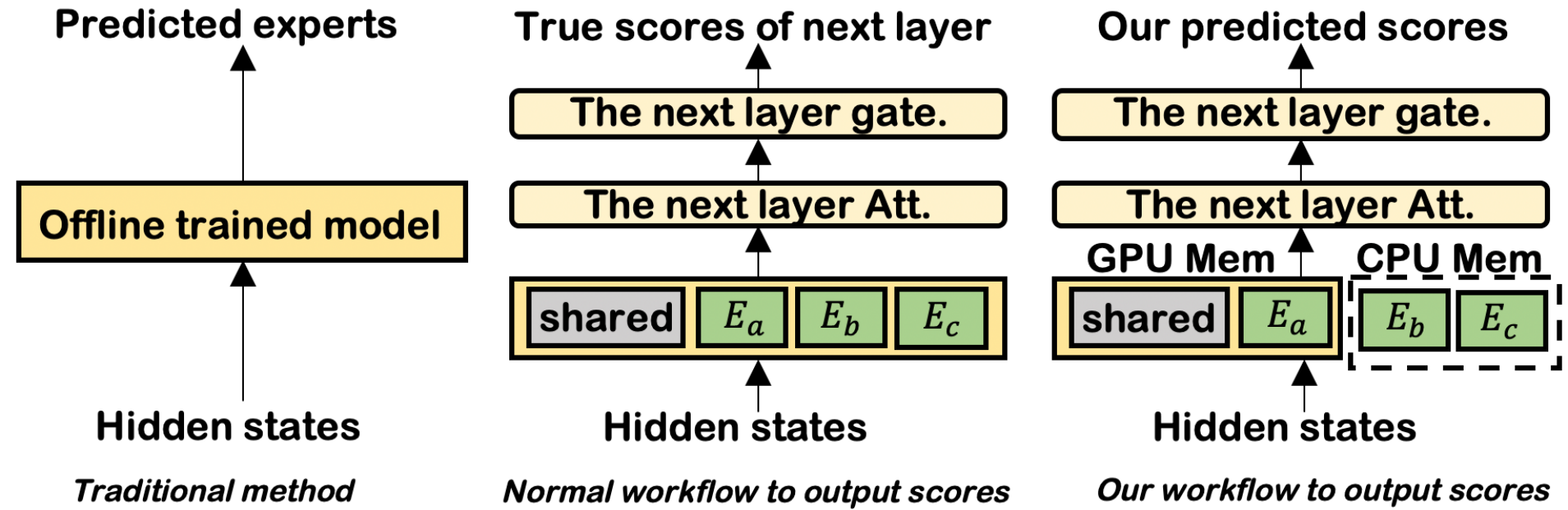}
        \caption{Our prefetching method compared to traditional method and normal workflow to output true scores.} \label{prefetch}
    \end{minipage}
\end{figure*}
\noindent
\textbf{Cache eviction.}\label{cacheeviction}
The cache eviction policy is instrumental in managing the removal of experts from GPU memory when new experts are loaded. 
As depicted in Fig. \ref{expertreuse}, experts that achieve higher scores, even if they are inactive in a particular iteration, have a greater inherent probability of being reused in the next 3 iterations, either as top-k or alternative experts, compared to those with minimal scores.
This indicates that experts with high scores in the current iteration have consistently achieved high scores in recent decoding iterations.
To preserve experts in the GPU with scores similar to those of active low-score experts, we employ a score-based strategy. This strategy involves evicting the expert that exhibits the lowest average activation score accumulated over the preceding \( n \) iterations. Specifically, for an expert \( i \), its score at the \( j \)-th token generation during decoding is denoted as \( S_{i,j} \). With \( m \) experts in total, the expert to be evicted is the one with the minimum average score calculated as follows:

\begin{equation}
\text{Evicted expert} = \arg\min_{i} \ \frac{\sum_{k=\max(1, j-n)}^{j} S_{i,k} }{j - \max(1, j-n) + 1} 
\end{equation}

This approach prioritizes the retention of experts that have demonstrated a higher historical impact, as measured by their contribution to model outputs, over a strategy that relies solely on the recency of access, such as Least Recently Used (LRU).
Crucially, our score-aware eviction policy takes into account the activation scores of all experts accessed within the observation window of \( n \) iterations. This includes inactive experts that can serve as substitutes for low-score experts, ensuring that the GPU retains a pool of experts that are both relevant and likely to enhance future computations.

Moreover, to prevent evicting an expert immediately before its use, the system dynamically elevates the eviction priority of any expert selected by the expert router for computation in a layer.
This temporary protection shield ensures the expert remains resident in GPU memory throughout its required computation window. 
The shield is automatically revoked upon completion of the layer's computation, returning the expert to standard eviction eligibility based on its score history.

\subsection{Loading Top-score Expert Online}
Our prefetching prioritizes top-score experts to meet \textbf{\textit{C$_4$}}. This technique effectively overlaps the latency of expert loading with ongoing computation time. Additionally, we aim to minimize the cost of prefetching to meet \textbf{\textit{C$_5$}} and enhance the accuracy of prefetching to meet \textbf{\textit{C$_6$}}.

As shown in Fig. \ref{prefetch}, the prediction process is carried out entirely with the parameters in GPU. 
Due to the GPU's superior speed, it tends to be more idle compared to the CPU and PCIe. 
As a result, the time taken by the GPU for prediction is often overshadowed by the time required for calculations by the CPU and loading by PCIe.
Consequently, the time cost of prediction on TPOT is rather minimal.

Moreover, we can ensure the accuracy of prefetching.
We perform calculations on both unshared experts currently in the GPU memory and shared experts, generating hidden states. These hidden states are then processed using the next layer's key-value cache to complete the attention computation. Subsequently, we carry out a gate computation to determine the scores for all experts in that layer.
The gate computation results, derived from calculations involving the shared experts and experts in GPU memory, are more accurate for two main reasons. First, shared experts process universal information across all inputs and remain constantly available in the GPU memory, which enhances the accuracy of the computations. Second, our cache eviction strategy, as detailed in Section \ref{cacheeviction}, ensures that high-scoring and thus important experts are retained in the cache during the most recent decoding stages. These factors collectively lead to score results that closely align with the true outcomes. If a prediction is wrong, the system simply reverts to the baseline state without performance penalties. We immediately clear mispredicted experts from the load queue and insert the correct ones.

While prior works like HybriMoE \cite{zhong2025hybrimoe} utilize prefetching, their approach predicts multiple future layers and attempts to prefetch all absent experts.
Because PCIe loading is significantly slower than computation, this untargeted approach creates excessive memory bandwidth pressure.
Consequently, the system often fails to prefetch all required experts in time, bottlenecking the pipeline.
In contrast, we classify experts into top-score and low-score via the gating network and strictly prefetch only the top-score experts.
This targeted approach drastically reduces PCIe load pressure, ensuring a realistic and effective overlap between compute and fetch times. 
Furthermore, predicting only a small subset of top-score experts inherently yields higher prediction accuracy because top-score experts remain high-impact on the output of the next layer even with incomplete expert computation.

\begin{figure}[b]

    \begin{minipage}{\linewidth}
        \centering
        \includegraphics[width=\linewidth]{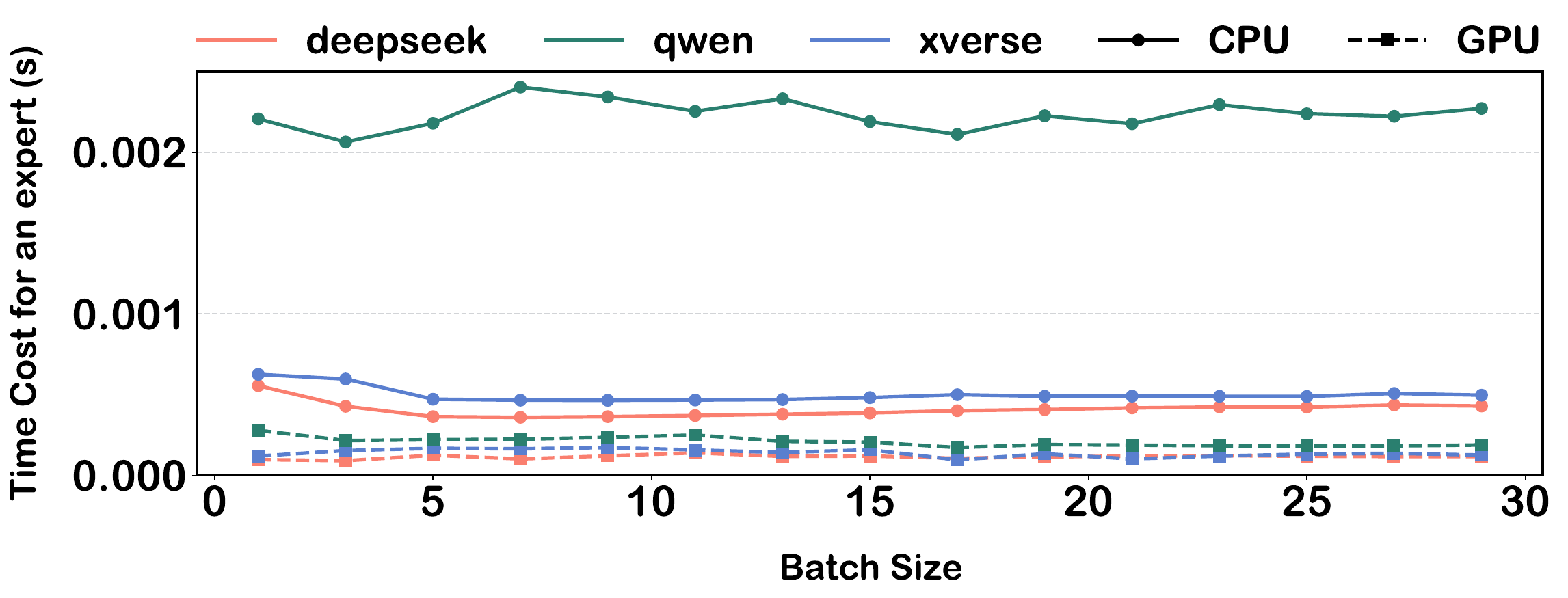}
        \caption{GPU (A6000) vs CPU (8-core) at low batch.}
        \label{GPUCPUtime}
    \end{minipage}
\end{figure} 

\begin{figure}[t]
\begin{algorithm}[H]
\caption{CPU-assisted Task Load Scheduling}
\label{alg:cpu_balance}
\begin{algorithmic}[1]
\Require $S$, $C_{CPU}$, $C_{load}$
\Ensure $n_{load}$, $n_{CPU}$
\State $S$: List of experts $\{uid_i\}$ sorted in descending order by score
\State $T_{CPU} \gets 0$, $T_{load} \gets 0$, $n_{load} \gets 0$, $n_{CPU} \gets 0$
\State $l \gets 0$, $r \gets |S| - 1$
\While{$l \leq r$}
  \If{$T_{load} \leq T_{CPU}$}
    \State $T_{load} \gets T_{load} + C_{load}$, $n_{load} \gets n_{load} + 1$
    \State $l \gets l + 1$
  \Else
    \State $T_{CPU} \gets T_{CPU} + C_{CPU}$, $n_{CPU} \gets n_{CPU} + 1$
    \State $r \gets r - 1$
  \EndIf
\EndWhile
\end{algorithmic}
\end{algorithm}
\end{figure}
\section{System Implementation}
This section details the implementation for efficient GPU-CPU coordination in expert-offloading.
Despite improvements in the GPU expert cache ratio via low-score substitution and top-score prefetching, some experts still need to be transferred from CPU to GPU due to prefetching failures from high demand or low PCIe bandwidth, and predictive errors. The expert-cache router may also fail to replace all low-score experts in GPU memory, particularly with significant score differences from inactive experts. 
To address these issues, we introduce CPU-assisted task load scheduling to balance CPU and GPU loads, boosting system efficiency. 
Additionally, we give a pipeline example between Layers during SMoE decoding with CPU-assisted computing to show how our design schedules tasks to reduce execution bubbles.
\subsection{CPU-assisted Task Load Scheduling}\label{management}

The  CPU-assisted task load scheduling system is designed to dynamically decide whether to transfer activated experts, which are not pre-cached in GPU memory, to the GPU via PCIe for computation or to perform the calculations directly on the CPU. This decision-making process is guided by Algorithm \ref{alg:cpu_balance}, which uses a two-pointer strategy to balance the cumulative costs associated with loading and CPU computation times. 
In addition to balancing these costs, we prioritize loading high-scoring experts into GPU memory because, as shown in Fig. \ref{expertreuse}, these experts are more likely to be reused.

The scheduling system aims to minimize idle times by balancing expert loading time, \( T_{load} = n_{load} \times C_{load} \), with CPU computation time, \( T_{CPU} = n_{CPU} \times C_{CPU} \). Here, \( C_{load} \) is the time to transfer an expert via PCIe, and \( C_{CPU} \) is the average time for a CPU computation. The system uses the past \( p \) instances of these times to optimize scheduling.
\( C_{CPU} \) is treated as a constant due to stable CPU times across low-batch sizes, as shown in Fig. \ref{GPUCPUtime}, simplifying our model.
Fig. \ref{GPUCPUload} indicates minimal GPU operation times compared to CPU and loading times, allowing us to approximate the total time by \( C_{load} \) alone. Thus, \( C_{load} \) is the primary cost factor, with \( C_{CPU} \) as a constant.
The optimization goal is expressed by the following equation, which seeks to minimize the maximum of \( T_{load} \) and \( T_{CPU} \):
\begin{equation}
 \min\ \max(n_{load} \times C_{load}, n_{CPU} \times C_{cost})
\end{equation}

Unlike frameworks that offload heavy computation to the CPU, SMoE’s substitution-centric design minimizes CPU involvement. This not only supports a wider range of legacy/edge CPUs lacking specialized instructions but also eliminates the sensitivity to batch-size fluctuations inherent in CPU-based computation.

\subsection{Pipeline Example between Layers}

\begin{figure}[t]

    \begin{minipage}{\linewidth}
        \centering
        \includegraphics[width=\linewidth]{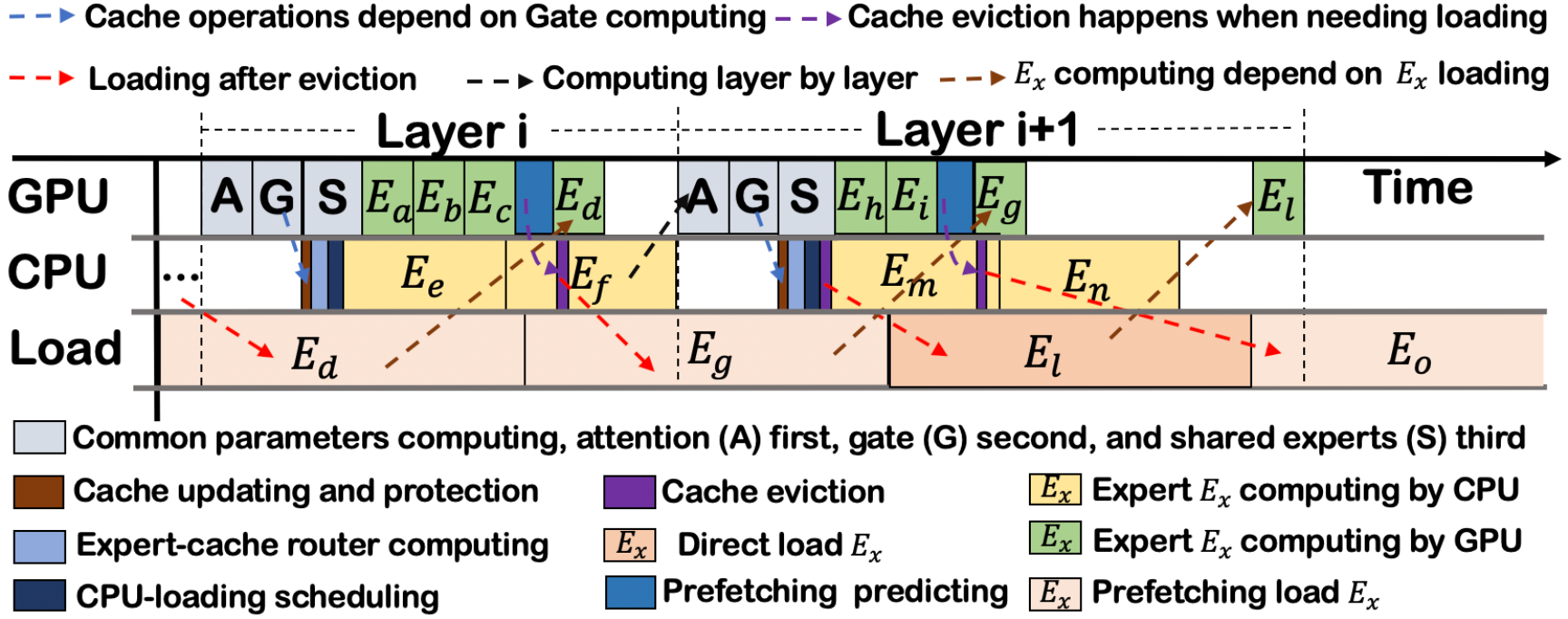}
        \caption{SMoE pipelines GPU, CPU, and load operations between two MoE layers.} \label{pipeline}
    \end{minipage}
\end{figure}

Fig. \ref{pipeline} illustrates the SMoE pipeline, which orchestrates GPU, CPU, and PCIe resources across consecutive layers ($i$ and $i+1$).
Colored blocks denote distinct operation phases. Dashed arrows explicitly indicate critical data dependencies, which include: (1) cache updates and protection shield activation are triggered strictly after Gating and attention; (2) GPU expert computation initiates only after full data loading; and (3) layer $i+1$ execution is serialized after layer $i$.
While these dependencies introduce execution bubbles, our pipeline design effectively masks these overheads within the dominant PCIe transfer latency.

\vspace{1mm}
\noindent
\textbf{CPU Operations:} The CPU operations are divided into four key parts, all integral to the SMoE's functionality and the computation of expert parameters on the CPU:

\vspace{1mm}
\noindent
(1) \textit{\textbf{Expert-cache router calculation:}} This part replaces some low-score experts from layer $i$ with inactive experts in GPU.

\vspace{1mm}
\noindent
(2) \textit{\textbf{CPU-assisted task load scheduling:}} This process determines which experts should be computed directly on the CPU and which should be loaded into the GPU.

\vspace{1mm}
\noindent
(3) \textit{\textbf{CPU expert computation:}} CPU computes the parameters of experts from layer $i$ not in GPU.

\noindent
(4) \textit{\textbf{Cache eviction and Protection Shield:}} Cache eviction decides which expert to evict when loading new experts into the GPU. Protection shield maintains the experts in the current layer to prevent evicting an expert immediately before its use. Both of these operations can be overlarpped by GPU computing and expert loading by PCIe.

\vspace{1mm}
\noindent
\textbf{GPU Operations:} The GPU operations are divided into four parts, three of which focus on computing expert parameters from layer $i$, while one is dedicated to prefetching parameters from layer $i+1$:

\vspace{1mm}
\noindent
(1) \textit{\textbf{Common parameter computation:}} This part computes the attention and gate parameters within the common parameters of layer $i$ on the GPU. 
%

\vspace{1mm}
\noindent
(2) \textit{\textbf{Direct expert computation:}} This part computes experts that are already in the GPU.

\vspace{1mm}
\noindent
(3) \textit{\textbf{Expert Prediction for Prefetching:}} This process predicts which experts from layer $i+1$ should be prefetched.

\vspace{1mm}
\noindent
(4) \textit{\textbf{New Expert Computation:}} This part continues the computation of newly loaded experts once the PCIe loading process is complete.

\vspace{1mm}
\noindent
\textbf{PCIe Load Operations:} The PCIe load operations are divided into two parts:

\vspace{1mm}
\noindent
(1) \textit{\textbf{Prefetching Experts from Layer $i+1$:}} This part prefetches the necessary experts from layer $i+1$ to the GPU in advance.

\vspace{1mm}
\noindent
(2) \textit{\textbf{Loading Experts from Layer $i$:}} This part loads the experts from layer $i$ into the GPU immediately as needed.



\subsection{Selecting value for hyperparameter $\alpha$}
\begin{figure}[t]

    \begin{minipage}{\linewidth}
        \centering
        \includegraphics[width=\linewidth]{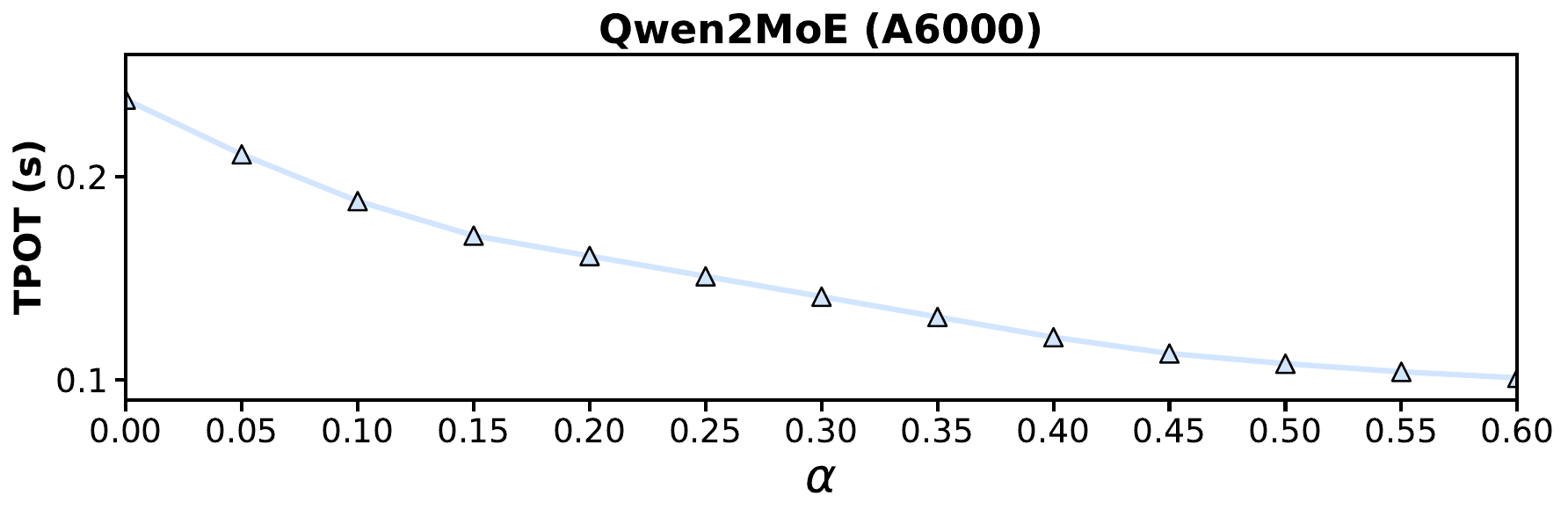}
        \caption{Relationship between $\alpha$ and $T(\alpha)$.} \label{imple}
    \end{minipage}
\end{figure}

Within our scheduler, the hyperparameter $\alpha$ defines the ratio of experts to be substituted, controlling the tradeoff between accuracy and token generation speed. 
An appropriate $\alpha$ identifies closely scored experts, including low-score and GPU-resident cached experts, that can be safely substituted with minimal accuracy loss. 
A larger $\alpha$ reduces latency by allowing more low-score experts to be replaced, but it may increase the score gap between substituted and cached experts, potentially amplifying output deviation. 
Besides, the $\alpha$ selection is also model/workload dependent.
To address this, we formalize it as the following constrained optimization problem:

\begin{equation}
    \min_{\alpha} A(\alpha) \quad \text{s.t.} \quad T(\alpha) \le R,
\end{equation}
where $A(\alpha)$ denotes the accuracy loss, $T(\alpha)$ is the average TPOT, and $R$ is the TPOT budget specified by the user.
Here, we assume that the user has a latency requirement while aiming to maximize accuracy.
To solve this, we first observe that the latency metric (or TPOT) generally decreases linearly with the value of $\alpha$. 
Intuitively, a larger $\alpha$ allows more low-score experts to be substituted, which increases the computation performed on the GPU and hence decreases the TPOT.
Based on this observation, as shown in Fig.~\ref{imple}, we can approximate $T(\alpha)$ by fitting a low-degree polynomial to the empirical relation between $\alpha$ and $T(\alpha)$. 
This yields a smooth and simple function that can be evaluated efficiently.
Given a target latency constraint $R$, we then perform a simple one-dimensional search over $\alpha$ and select the smallest value whose predicted $T(\alpha)$ satisfies the constraint, thereby ensuring minimal accuracy loss while meeting the latency requirement.

\begin{table}[t]
\caption{Model and GPU Configurations}
\centering
\setlength{\tabcolsep}{3pt}
\begin{tabular}{ccc}
\toprule
Setting & Model & GPU \\
\midrule
S1 & deepseek-moe-16b \cite{deepseekmoe} (31GB)& 3080ti (12GB) \\
S2 & XVERSE-MoE-A4.2B \cite{xversemoe} (49GB) & 4060ti (16GB) \\
S3 & Qwen2-57B-A14B \cite{qwenmoe} (107GB) & A6000 (48GB)\\
\bottomrule
\end{tabular}
\label{settings}
\end{table}
\begin{figure*}[t]
    \begin{minipage}{\linewidth}
        \centering
        \includegraphics[width=\linewidth]{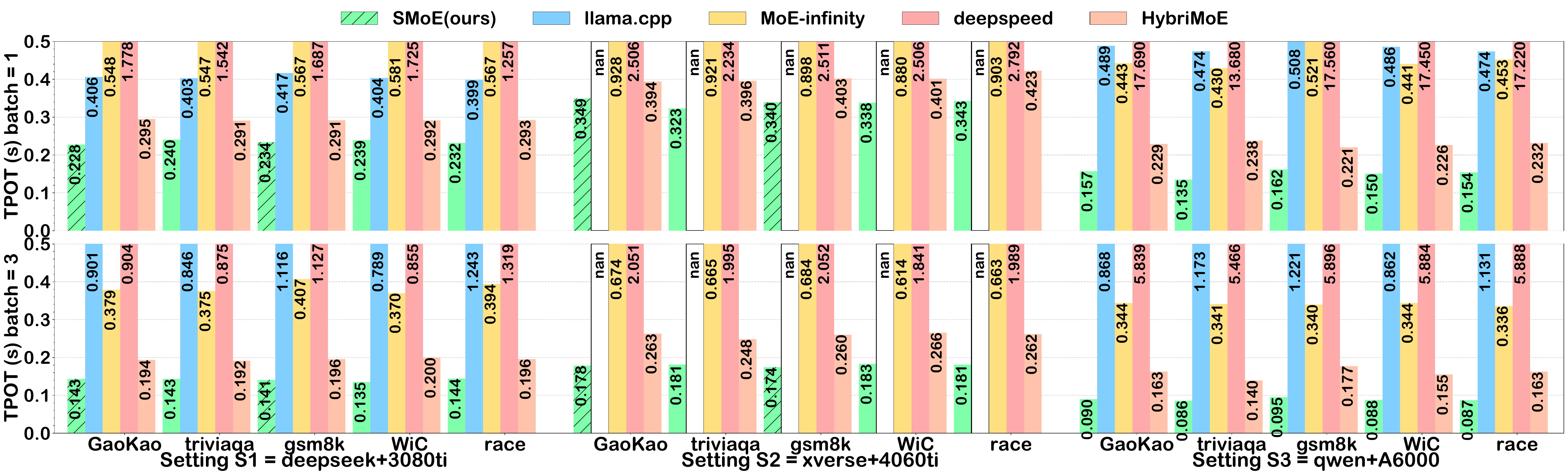}
        \caption{TPOT of four baselines and our method in five workloads.} \label{overallTPOT}
    \end{minipage}
\end{figure*}
\begin{table*}[t]
\centering
\caption{Generation quality across workloads and models at various expert substitution thresholds $\alpha$.}
\label{accuracy}
\begin{tabular}{lcccccccccccccc} 
\toprule
Workloads & Model & 0.0 & 0.05 & 0.1 & 0.15 & 0.2 & 0.25 & 0.3 & 0.35 & 0.4 & 0.45 & 0.5 & 0.55 & 0.6 \\
\midrule
\multirow{3}{*}{\makecell[l]{GaoKao\\Acc. (\%)}} & Deepseek & 27.2 & 27.6 & 28.2 & 29.5 & 29.3 & 28.9 & 28.1 & 27.7 & 27.3 & 26.8 & 26.7 & 26.2 & 25.9 \\
& Xverse & 47.2 & 47.5 & 47.9 & 48.1 & 49.0 & 47.2 & 47.9 & 47.5 & 46.5 & 46.8 & 47.5 & 46.2 & 45.9 \\
& Qwen & 73.5 & 74.2 &74.8 & 75.8 & 76.0 & 76.1 & 74.2 & 74.1 & 73.2 & 72.1 & 71.6 & 70.7 & 71.7 \\
\midrule
\multirow{3}{*}{\makecell[l]{WiC\\Acc. (\%)}} & Deepseek & 50.7 & 50.9 & 51.3 & 51.6 & 50.9 & 51.7 & 50.6 & 50.7 & 51.9 & 51.8 & 51.7 & 50.1 & 50.4 \\
& Xverse & 50.0 & 50.1 & 50.3 & 50.2 & 50.2 & 50.0 & 50.0 & 50.2 & 50.2 & 50.1 & 50.00 & 49.8 & 49.8 \\
& Qwen & 60.5 & 60.7 & 60.5 & 60.6 & 60.9 & 60.7 & 60.6 & 60.7 & 61.9 & 60.8 & 60.3 & 60.4 & 60.1 \\
\midrule
\multirow{3}{*}{\makecell[l]{Triviaqa\\Acc. (\%)}} & Deepseek & 59.3 & 59.1 & 59.3 & 58.5 & 58.5 & 57.7 & 59.6 & 59.0 & 58.9 & 58.7 & 58.6 & 57.4 & 57.6 \\
& Xverse & 53.1 & 53.2 & 52.8 & 52.7 & 53.1 & 53.3 & 53.4 & 53.8 & 52.8 & 53.1 & 53.2 & 52.1 & 52.4 \\
& Qwen & 69.3 & 69.2 & 68.7 & 68.5 & 68.5 & 67.7 & 69.6 & 69.0 & 68.9 & 68.7 & 68.6 & 67.6 & 67.9 \\
\midrule
\multirow{3}{*}{\makecell[l]{Race-mid\\Acc. (\%)}} & Deepseek & 70.0 & 69.8 & 69.2 & 69.7 & 69.9 & 70.4 & 70.0 & 69.1 & 70.2 & 70.2 & 70.4 & 68.9 & 68.6 \\
& Xverse & 81.4 & 81.6 & 81.8 & 81.0 & 82.3 & 82.0 & 81.9 & 80.9 & 82.9 & 82.3 & 82.4 & 81.1 & 80.3 \\
& Qwen & 80.0 & 79.8 & 79.9 & 79.7 & 79.9 & 80.4 & 80.0 & 79.1 & 80.2 & 80.1 & 80.4 & 79.8 & 79.6 \\
\midrule
\multirow{3}{*}{\makecell[l]{Gsm8k\\Acc. (\%)}} & Deepseek & 51.4 & 51.6 & 51.2 & 51.4 & 49.5 & 51.9 & 51.0 & 50.2 & 49.2 & 48.8 & 47.8 & 47.9 & 48.8 \\
& Xverse & 62.9 & 62.7 & 62.9 & 62.4 & 63.3 & 61.7 & 62.5 & 61.2 & 61.2 & 60.3 & 60.4 & 60.1 & 58.6 \\
& Qwen & 85.7 &85.5 & 85.7 & 85.2 & 85.8 & 85.6 & 85.1 & 85.3 & 84.5 & 84.1 & 83.7 & 83.2 & 83.3 \\
\midrule

\multirow{3}{*}{\makecell[l]{MT-bench\\Score (1-10)}} & Deepseek & 1.43 & 1.42 & 1.58 & 1.53 & 1.59 & 1.47 & 1.39 & 1.41 & 1.43 & 1.39 & 1.45 &1.20 & 1.50 \\
& Xverse & 6.41 & 6.30 & 6.17 & 5.91 & 6.17 & 6.08 & 5.79 & 6.11 & 6.15 & 5.81 & 5.50 & 5.61& 5.70\\
& Qwen & 8.52 &8.43 & 8.39 & 8.39 & 8.40 & 8.37 & 8.33 & 8.21 & 8.28 & 8.29 & 8.36 & 8.23 & 8.19 \\
\midrule
\multirow{3}{*}{\makecell[l]{MMLU\\Acc. (\%)}} & Deepseek & 38.0 & 39.1& 39.8 & 39.2 & 39.4 & 39.4 & 39.8 & 38.9 & 39.2 & 39.1 & 39.2 & 37.2 & 36.1 \\
& Xverse & 39.4 & 40.3 & 43.1 & 37.8 & 42.3 & 41.9 & 42.3 & 41.7 & 42.1 & 43.9 & 45.2 & 46.3 & 42.9 \\
& Qwen &71.0 &70.8 & 70.0 & 70.3& 70.6 & 71.0 & 69.8& 68.8& 71.0 &67.8 & 71.0 &71.0 & 70.4 \\
\midrule
\multirow{3}{*}{\makecell[l]{HumanEval\\pass@1(\%)}} & Deepseek & 26.7 & 26.5 & 27.4 & 26.3 & 25.0 & 24.1 & 26.2 & 22.7 & 23.7 & 22.4& 21.3 & 21.6& 20.7\\
& Xverse & 46.3 & 48.7 & 51.2 & 49.6 & 48.2 & 47.3 & 52.4 & 51.9 & 50.0 & 49.7 & 52.4 & 48.3 & 47.6 \\
& Qwen & 53.0 &51.8 & 51.2 & 50.7 & 50.0 & 53.3 & 56.7 & 53.4&50.6 & 50.9 & 50.6 & 48.7 & 47.6 \\
\bottomrule
\end{tabular}
\label{tab:accuracy_comparison}
\end{table*}

\section{Evaluation}
This section aims to demonstrate that our method, SMoE, significantly reduces the latency of token generation during the decoding phase, increases the hit rate of experts within the GPU, and causes almost no loss in accuracy. 
Additionally, we conduct experiments to illustrate the impact of various components (expert-cache router, cache eviction strategy, prefetching, and CPU-assisted task load scheduling) on decoding performance and to analyze the reasons behind these impacts.


\subsection{Setup}
In designing our experimental setup for LLM inference on edge devices, we focus on three key aspects. First, we select models exceeding edge GPU memory capacity to evaluate performance under typical resource constraints, reflecting real-world scenarios \cite{chkirbene2024large}. Second, we employ popular models with a fine-grained MoE architecture, effective as shown in recent studies \cite{deepseekmoe}. Third, we test diverse model types and GPU configurations to ensure result robustness and reliability.
We assess three experimental settings, detailed in Table \ref{settings}. Additionally, since MoE expert selection depends on workload, we test various workload types to show our method enhances inference speed without significant accuracy loss. Finally, we demonstrate our approach's superiority by comparing it with both the most advanced and popular methods, using metrics such as TPOT, TTFT, accuracy, and GPU expert cache ratio.


\vspace{1mm}
\noindent
\textbf{Models.}
We assess the performance of three widely recognized MoE models featuring a fine-grained MoE architecture: deepseek-moe-16b-base \cite{deepseekmoe}, Qwen2-57B-A14B-Instruct \cite{qwenmoe}, and XVERSE-MoE-A4.2B-Chat \cite{xversemoe}.
Given that the objective of this study is to facilitate lossless models, we do not run quantized models. 
However, it is worth noting that our approach is orthogonal to quantization operations, and thus, it can also be applied to serve quantized models effectively.

\vspace{1mm}
\noindent
\textbf{Hardware.} 
We test on a single NVIDIA RTX 3080 Ti GPU (12GB), a single NVIDIA RTX 4060 Ti GPU (24GB), and a single NVIDIA A6000 GPU (48GB).
We evaluated SMoE on two setups: (1) Edge/Legacy: RTX 3080Ti / 4060 Ti on PCIe 3.0 with Intel E5-2683 v3, and (2) High-end: NVIDIA A6000 on PCIe 4.0 with Intel Xeon Gold 6444Y.

\vspace{1mm}
\noindent
\textbf{Workloads.} 
We select these datasets to guarantee high diversity across evaluation tasks and knowledge domains. Gaokao~\cite{allenai:arc} and MMLU~\cite{hendryckstest2021} cover structured academic knowledge from secondary to college levels, including mathematics, science, social science and computer science. TriviaQA~\cite{2017arXivtriviaqa} and RACE-mid~\cite{lai-etal-2017-race} test open-domain QA and reading comprehension. WiC~\cite{pilehvar2018wic} evaluates contextual semantic understanding. GSM8K~\cite{Bisk2020} measures multi-step mathematical reasoning. MT-Bench~\cite{zheng2023judging} assesses conversational quality and instruction following. HumanEval~\cite{chen2021codex} evaluates code generation capability. This broad coverage enables a comprehensive and robust evaluation of our model.
We use the Math\_I, Math\_II, History, and Biology datasets in the Gaokao benchmark \cite{allenai:arc} and College\_computer\_science, Management, International\_law and Logical\_fallacies in the MMLU \cite{hendryckstest2021}.
%

\vspace{1mm}
\noindent
\textbf{Metrics.} We use three metrics to test inference speed: TPOT, TTFT and GPU cache ratio.
For decoding performance, we evaluate the TPOT as the key metric for the decoding stage. For prefilling performance, we assess the TTFT for the prefilling stage.
\begin{figure}[!b]
    \centering
    \includegraphics[width=\linewidth]{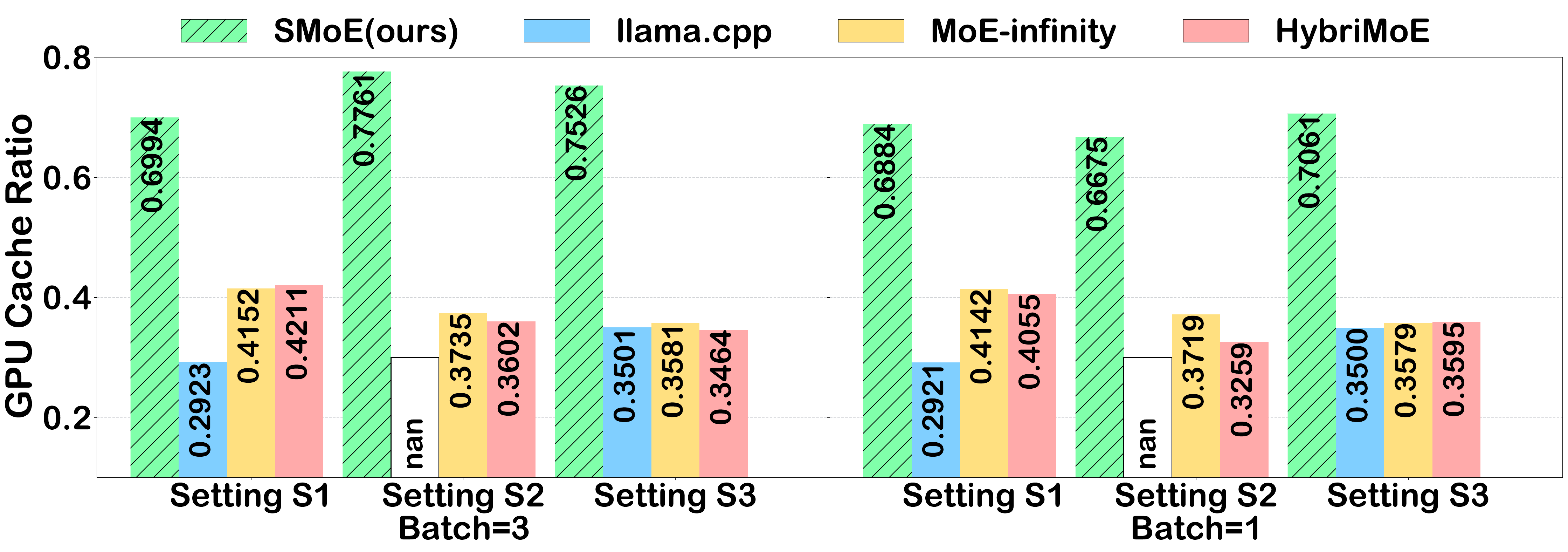}
    \caption{GPU cache ratio on average.}
    \label{cacheratio}
\end{figure}
\begin{figure*}[t]
    \begin{minipage}{0.47\linewidth}
        \centering
        \includegraphics[width=\linewidth]{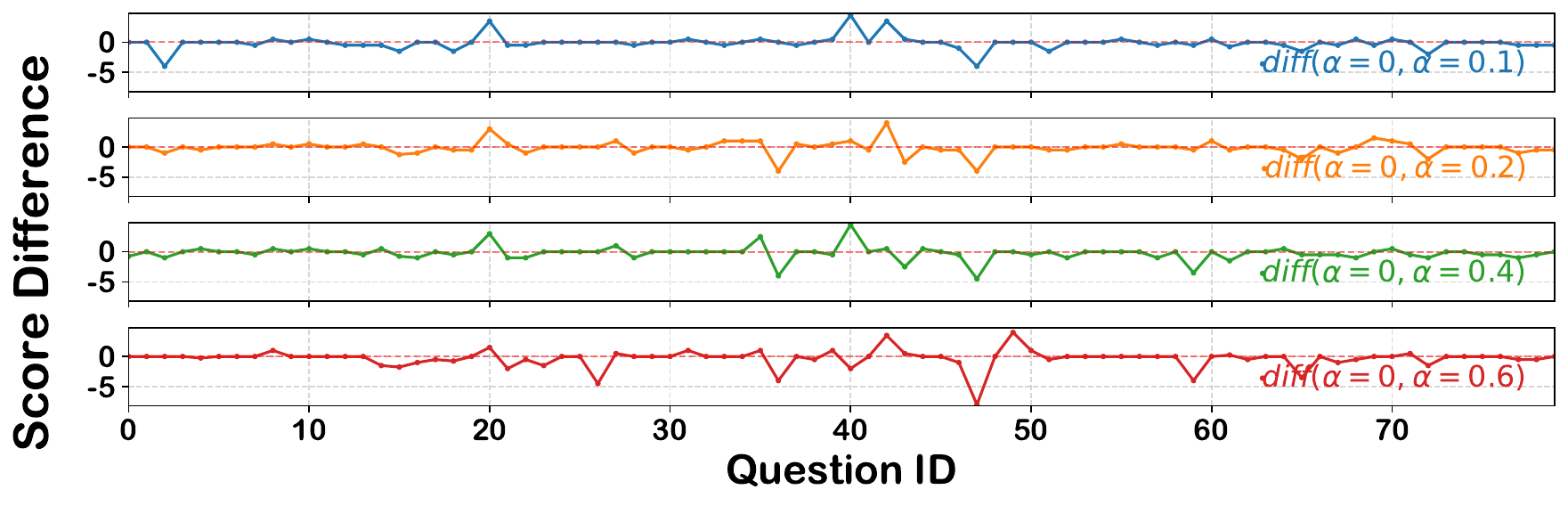}
        \caption{Correlation between ($\alpha$) and MT-bench score.} \label{mtbench_scores_diff}
    \end{minipage}
        \begin{minipage}{0.47\linewidth}
        \centering
        \includegraphics[width=\linewidth]{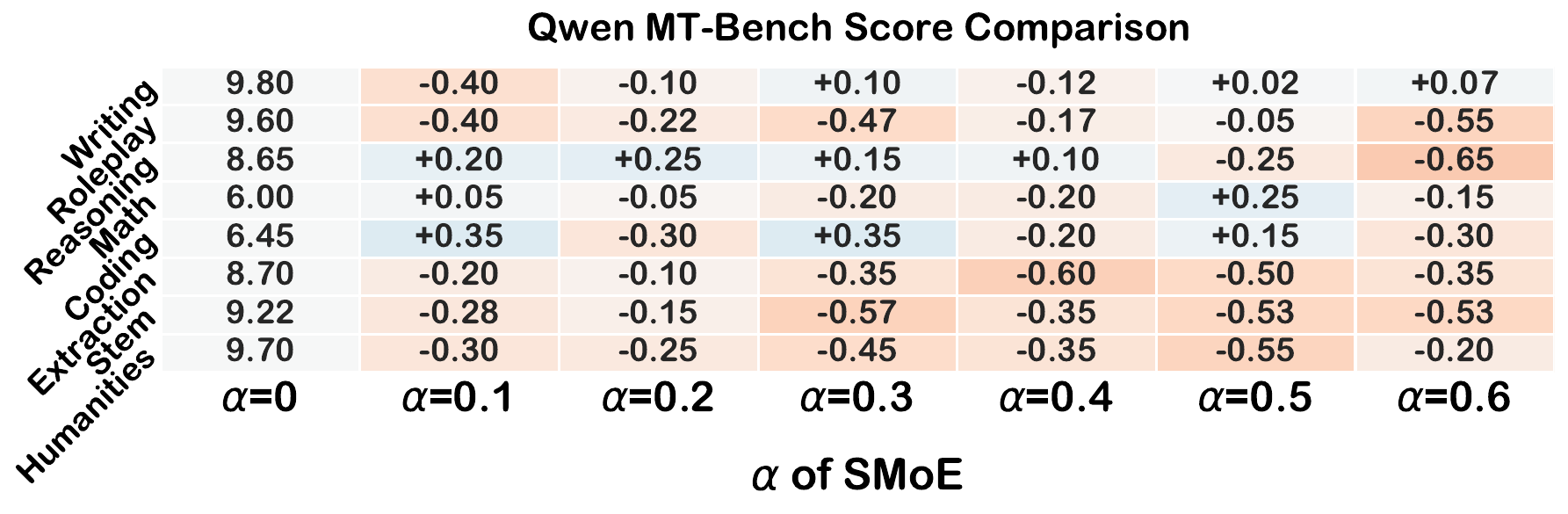}
    \caption{Per-domain Analysis of SMoE.}
    \label{hot}
    \end{minipage}
\end{figure*}
Additionally, we use the GPU expert cache ratio to reflect the GPU memory utilization efficiency, which is determined by the expert scheduling strategies of different methods.
We use three metrics to test model performance: Accuracy, GPT-4 Score, and pass@1. Specifically, OpenCompass \cite{opencompass} serves as a comprehensive framework to assess accuracy across diverse datasets, including Gaokao \cite{allenai:arc}, triviaQA \cite{2017arXivtriviaqa}, WiC \cite{pilehvar2018wic}, Race-mid \cite{lai-etal-2017-race}, gsm8k \cite{Bisk2020}, and MMLU \cite{hendryckstest2021}. For code generation capabilities, we evaluate the HumanEval dataset \cite{chen2021codex} using the pass@1 metric, which measures the percentage of problems solved correctly on the first attempt. Additionally, MT-Bench (GPT-4 Score) is designed to evaluate the multi-turn conversational and complex instruction-following abilities of LLMs through challenging open-ended prompts.
This is to demonstrate that the impact of our strategy on accuracy is minimal.

\noindent
\textbf{Baselines.} We evaluate our work, comparing it against baseline systems that support running MoE LLMs without enough GPU memory on the local platform: (1) \textit{\textbf{MoE-infinity}} \cite{xue2024moe} is designed for efficient MoE inference on personal machines with limited GPU memory. It leverages the high activation sparsity in the decode phase, where few experts are reused often. 
(2)\textit{\textbf{ Llama.cpp}} \cite{llamacpp} is a C++ implementation enabling efficient LLM inference on CPUs, optimized for low GPU memory devices.
(3) \textit{\textbf{DeepSpeed}} \cite{aminabadi2022deepspeed} optimizes large model training and inference by loading transformer layers onto the GPU layer-wise, enabling it to manage GPU-memory constraints without storing the entire model on the GPU.
(4) \textit{\textbf{HybriMoE}} \cite{zhong2025hybrimoe} is an offloading system designed to enhance resource utilization via a novel CPU-GPU scheduling and cache management framework. 
%
As HybriMoE is built on the ktransformer architecture \cite{ktransformer} with quantization, we've modified our implementation to remove these effects for fair evaluation comparisons.

\subsection{Overall Performance}
The overall performance section highlights our method's superior decoding performance compared to baselines, particularly at low batch sizes of 3 and 1, typical for edge deployment. 
We assess TTFT for prefilling performance at batch=1, demonstrating that our method matches or exceeds other approaches during the prefilling stage.
To reflect real-world workload variability, we sample about 1000 data points from each of five datasets and use basic prompts from openCompass \cite{opencompass}. 
Note that due to the Xverse model's incompatibility with the GGUF format, llama.cpp cannot run the S2 Xverse model, so we use nan to denote its value in our experiments.
%

\vspace{1mm}
\noindent
\textbf{Decoding performance.} Our method consistently achieves a 24\% reduction in TPOT compared to the best existing approach on average at batch=1, and a 35\% reduction at batch=3, as shown in Fig. \ref{overallTPOT}.
Notably, our method performs better when the batch size is 3 than when it is 1.  
This can be attributed to two key factors. 
First, in multi-batch scenarios, GPU computation time does not change significantly as the batch size increases, as shown in Fig. \ref{GPUCPUtime}.
Moreover, since more experts need to be loaded when decoding multiple tokens at a time, the proportion of PCIe load in TPOT increases.
Consequently, our strategies aimed at improving expert utilization in GPUs and reducing PCIe load become more effective.
Second, we can replace the low-score experts of a certain token with the top-score experts of other tokens in the same batch. 
Since top-score experts are inevitably computed, this approach avoids the need to additionally load the low-score experts, thus enhancing TPOT performance when batch=3.

Notably, our method significantly reduces TPOT under S3 settings, achieving a 48\% reduction on average at batch=3 compared to the best baseline, and 34\% reduction at batch=1. This is because the A6000 GPU boasts high computational speed, so TPOT in these settings is dominated by PCIe load time (in the case of MoE-infinity and HybriMoE) and CPU computation time (in the case of llama.cpp, DeepSpeed and HybriMoE). By improving expert utilization in GPU memory, our approach effectively reduces such CPU computations and PCIe loads.  
In contrast, performance gains in S1 are less pronounced, with a 27\% reduction on average at batch=3 compared to the best baseline, and a 20\% reduction at batch=1. 
This is because the computational capability of the 3080Ti GPU offers no significant advantage over CPU, which limits the benefits of improved GPU memory efficiency.
%

\vspace{1mm}
\noindent
\textbf{GPU expert cache ratio performance.}
As shown in Fig. \ref{cacheratio}, our method consistently achieves at least a 65\% improvement in cache ratio across all tested settings compared to the best existing approach. 
This confirms that our method directly enhances the utilization of experts in GPU memory.
In contrast, methods like llama.cpp and deepspeed implement offloading but are not designed for dynamic workloads, leading to static expert caching in the GPU that fails to adapt to workload changes. Due to deepspeed's sequential loading of layers into the GPU for computation, we don't conduct cache ratio experiments on it.
MoE-infinity \cite{xue2024moe} employs a prefetching strategy that can increase the hit rate of experts in the GPU based on the current workload, but it requires historical router data, which can lead to decreased prefetching accuracy when records are incomplete.
Moreover, prefetching alone is insufficient to load all active experts.

\vspace{1mm}
\noindent
\textbf{Accuracy performance.}
Evaluations conducted with our benchmarks, across diverse domain datasets show that our low-score expert substitution method causes only negligible accuracy variation when the substitution threshold is below 0.35, and even improves accuracy in some cases, as shown in Table \ref{accuracy}.
This occurs because, in MoE models, top-score experts are highly influential, while fluctuations in low-score expert contributions have little effect on the final output as long as a sufficient number of experts are retained.
To further enhance inference efficiency on edge devices, we favor a higher substitution ratio while maintaining this minimal loss.
This strategy classifies more experts as low-score, increasing substitution opportunities and reducing expensive PCIe data transfers.
In practice, we set the substitution threshold to 0.35 in S1, 0.3 in S2, and 0.25 in S3.

\begin{table}[b]
\centering
\caption{Performance comparison of Qwen2MoE.}
\label{qwen_comparison}
\setlength{\tabcolsep}{2pt}
\begin{tabular}{lcccc}
\toprule
\multirow{2}{*}{Benchmarks} & \multicolumn{4}{c}{Methods} \\
\cmidrule(lr){2-5}
 & Original & Skipping & Quantization & SMoE($\alpha=0.25$) \\
\midrule
GaoKao Acc. (\%) & 73.5 & 69.1 & 58.9& 76.1 \\
WiC Acc. (\%) & 60.5 & 60.66 & 59.8 & 60.7 \\
triviaqa Acc. (\%) & 69.3 & 67.9 & 62.1 & 67.7 \\
Race-mid Acc. (\%)& 80.0 & 80.0 & 78.9 & 80.4 \\
gsmk Acc. (\%)& 85.7 & 83.6 & 81.7 & 85.6 \\
MTbench (1-10)& 8.52 & 7.40 & 8.01 & 8.37 \\
MMLU & 71.0 & 70.2 & 51.1 & 71.0 \\
HumanEval pass@1(\%) & 53.0 & 53.0 & 31.1 & 53.3 \\
\bottomrule
\end{tabular}
\end{table}

As Table \ref{qwen_comparison} shows, SMoE maintains the highest fidelity, performing closest to the original model. The expert-skipping method \cite{lu2024not} yields the second-best performance, whereas quantization experiences a significant accuracy degradation. Specifically, the expert-skipping baseline directly discards the experts that would otherwise be substituted in SMoE. Consequently, the number of activated experts per token varies dynamically, with higher substitution rates leading to more dropped experts. Meanwhile, the GPTQ baseline quantizes a subset of layers to INT8, which is deliberately configured to achieve a GPU cache hit rate comparable to that of SMoE for a fair comparison.

Fig. \ref{mtbench_scores_diff} demonstrates that a higher substitution rate correlates with increased accuracy variance.
The plotted line tracks the difference in MT-Bench scores between SMoE (at the current $\alpha$) and the original qwenmoe, where larger fluctuations signify greater deviation from the baseline. The maximum variance is concentrated in questions 40--50, which correspond to the coding domain. Despite this, Fig. \ref{hot} confirms that our method achieves highly competitive performance across all domains, with only a slight performance penalty.

%

\begin{figure}[t]
    \begin{minipage}{\linewidth}
        \centering
        \includegraphics[width=\linewidth]{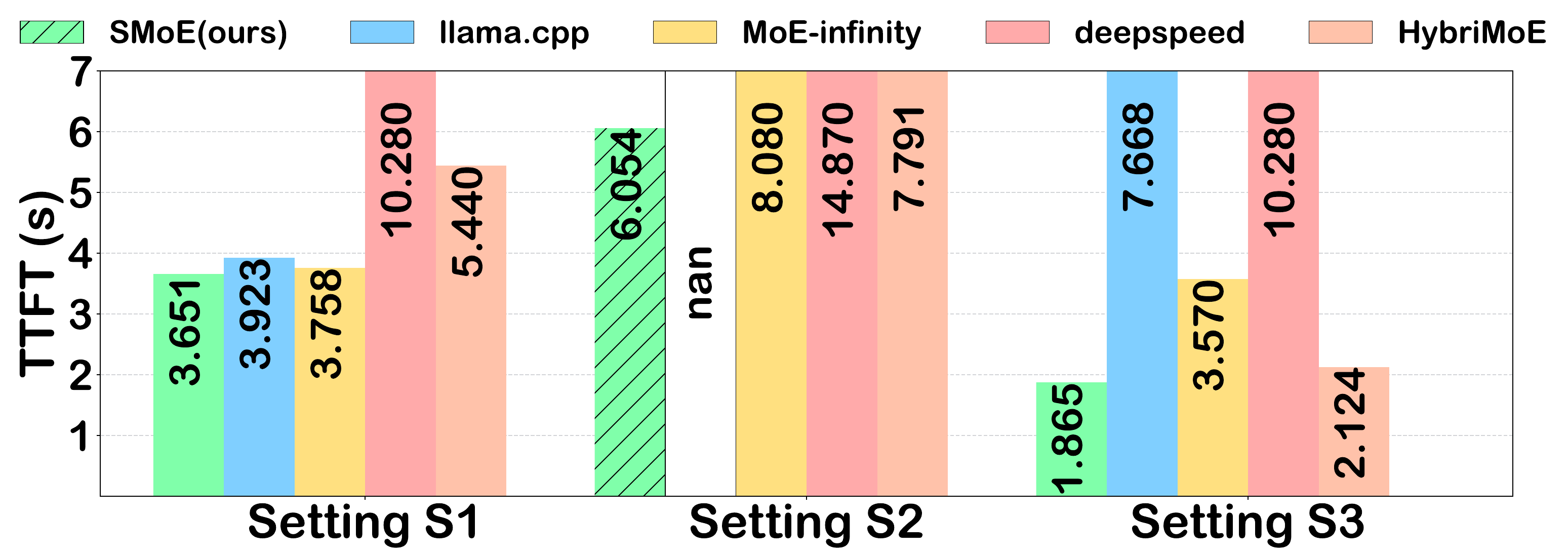}
        \caption{Prefilling time of baselines and SMoE on average.} \label{prefillingtime}
    \end{minipage}
\end{figure}
\begin{figure}
        \begin{minipage}{\linewidth}
        \centering
        \includegraphics[width=\linewidth]{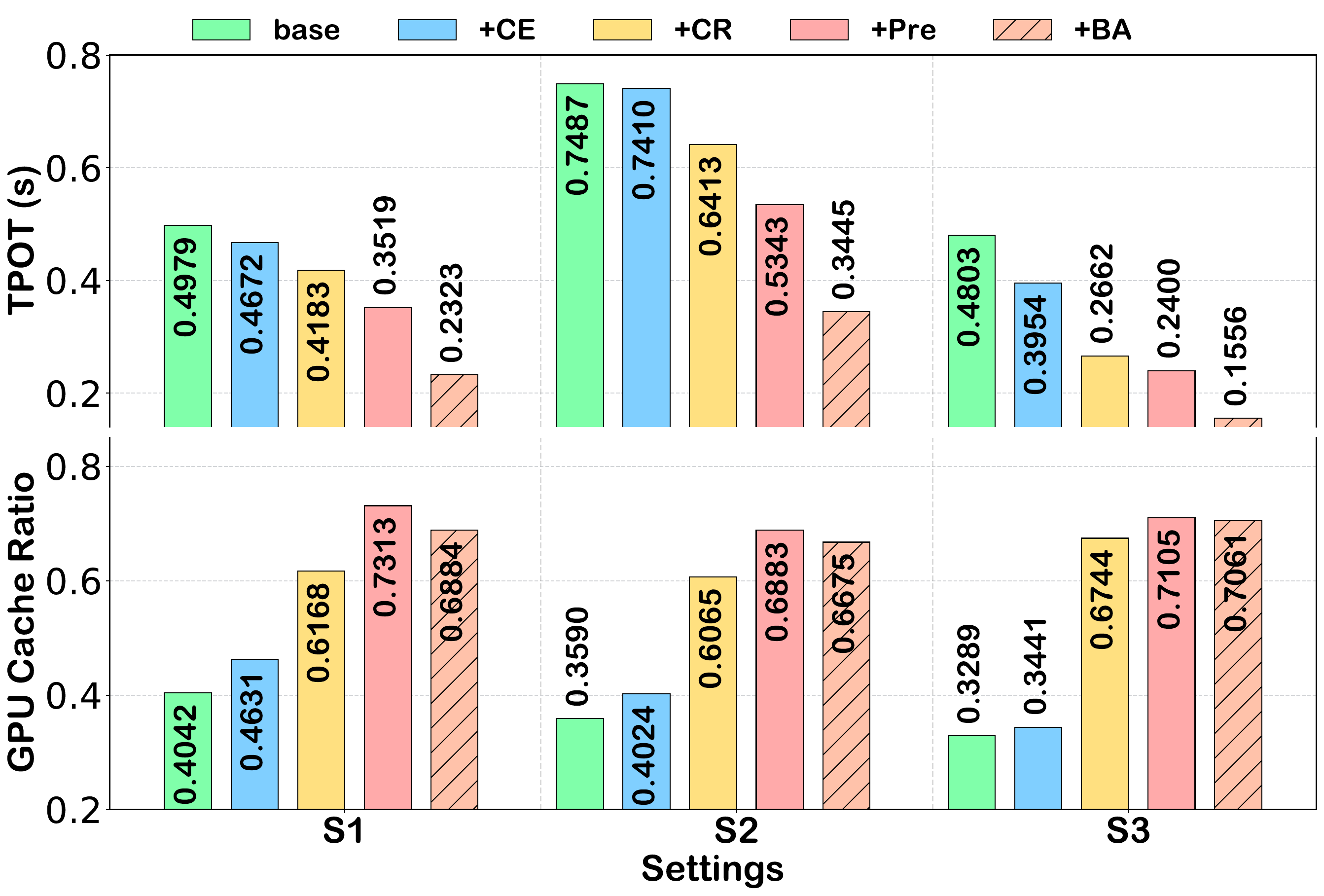}
    \caption{Impact of components on TPOT and cache ratio.}
    \label{breakdown}
    \end{minipage}
\end{figure}

\begin{figure}[b]
    \begin{minipage}{\linewidth}
        \centering
        \includegraphics[width=\linewidth]{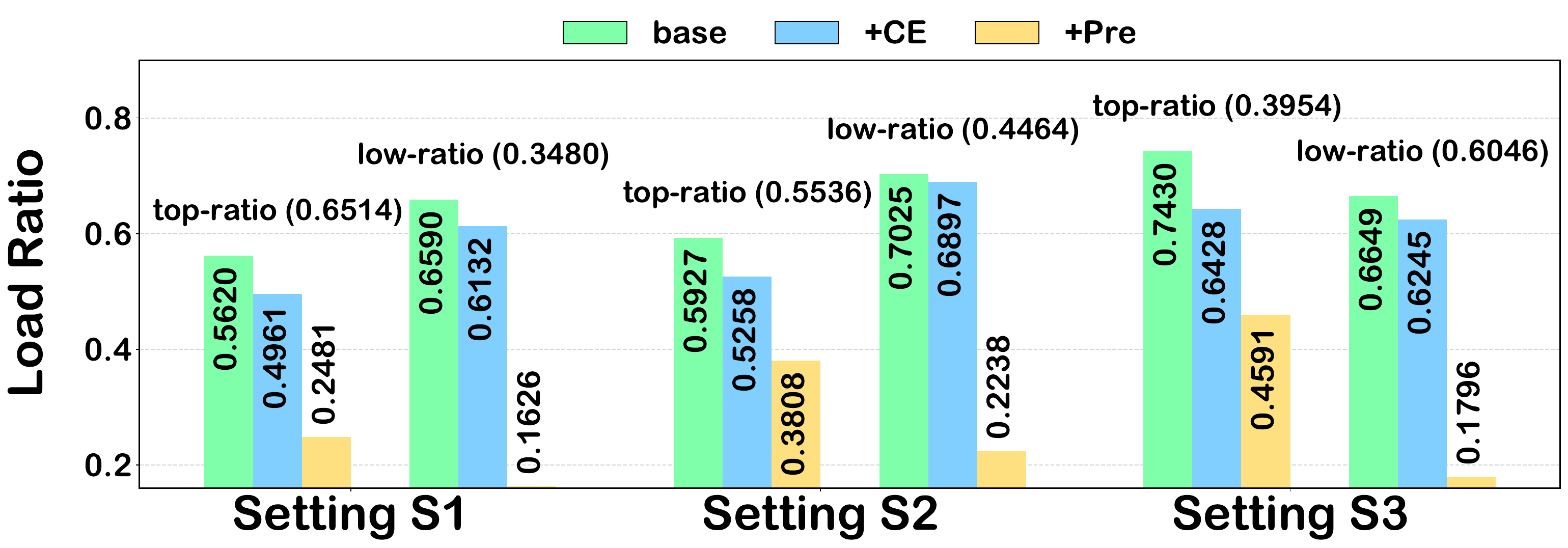}
        \caption{ \small Fewer top/low-score experts loading.}
\label{loadratio}
    \end{minipage}
\end{figure}
\begin{figure}[b]
    \begin{minipage}{\linewidth}
        \centering
        \includegraphics[width=\linewidth]{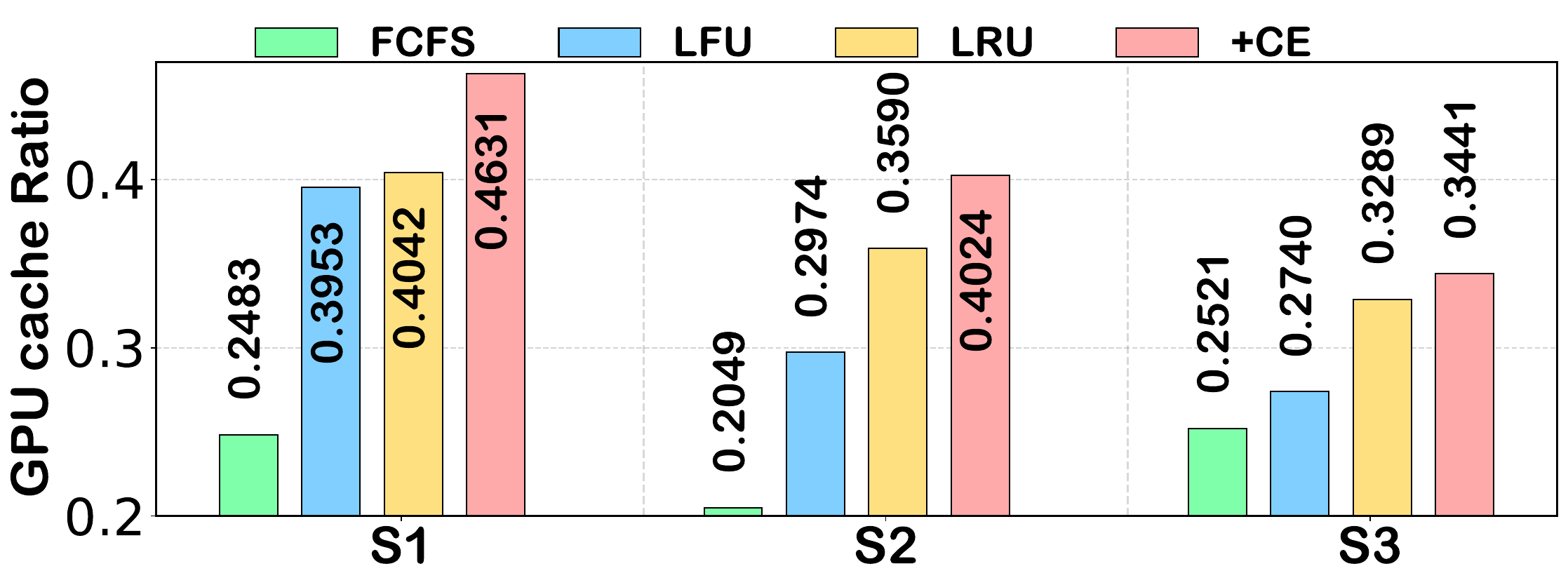}
        \caption{ \small "+CE" vs traditional cache methods.}
\label{cachecompare}
    \end{minipage}
\end{figure}

\begin{figure*}
    \begin{minipage}{0.33\linewidth}
        \centering
        \includegraphics[width=\linewidth]{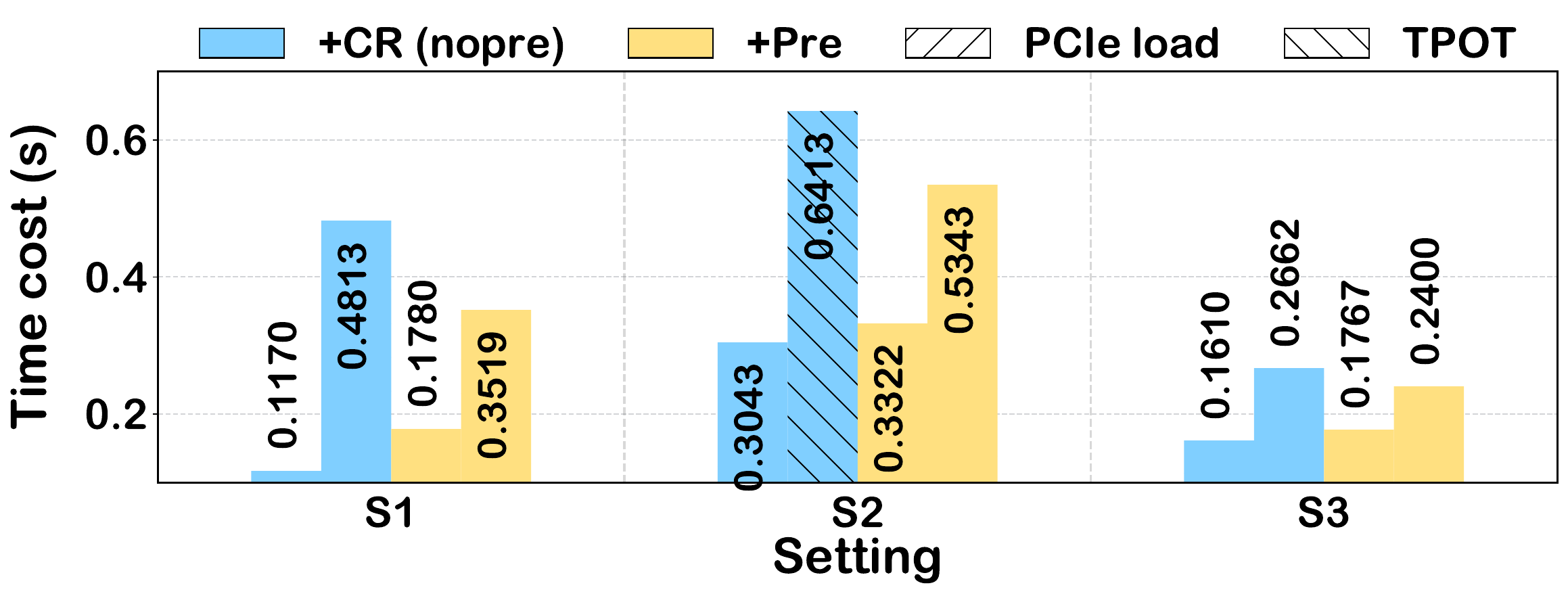}
        \caption{PCIe time vs TPOT.}
\label{pciechange}
    \end{minipage}
    \begin{minipage}{0.33\linewidth}
        \centering
        \includegraphics[width=\linewidth]{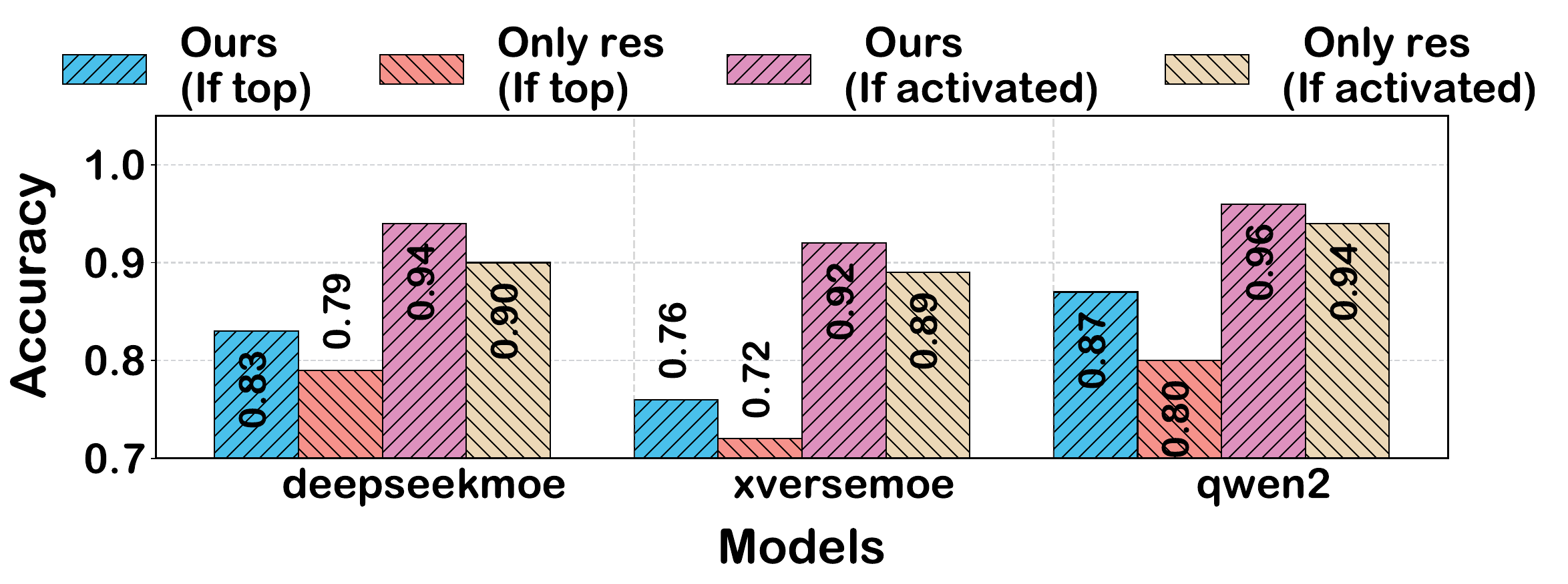}
        \caption{Prefetching accuracy.}
\label{prefetchaccuracy}
    \end{minipage}
        \begin{minipage}{0.33\linewidth}
        \centering
        \includegraphics[width=\linewidth]{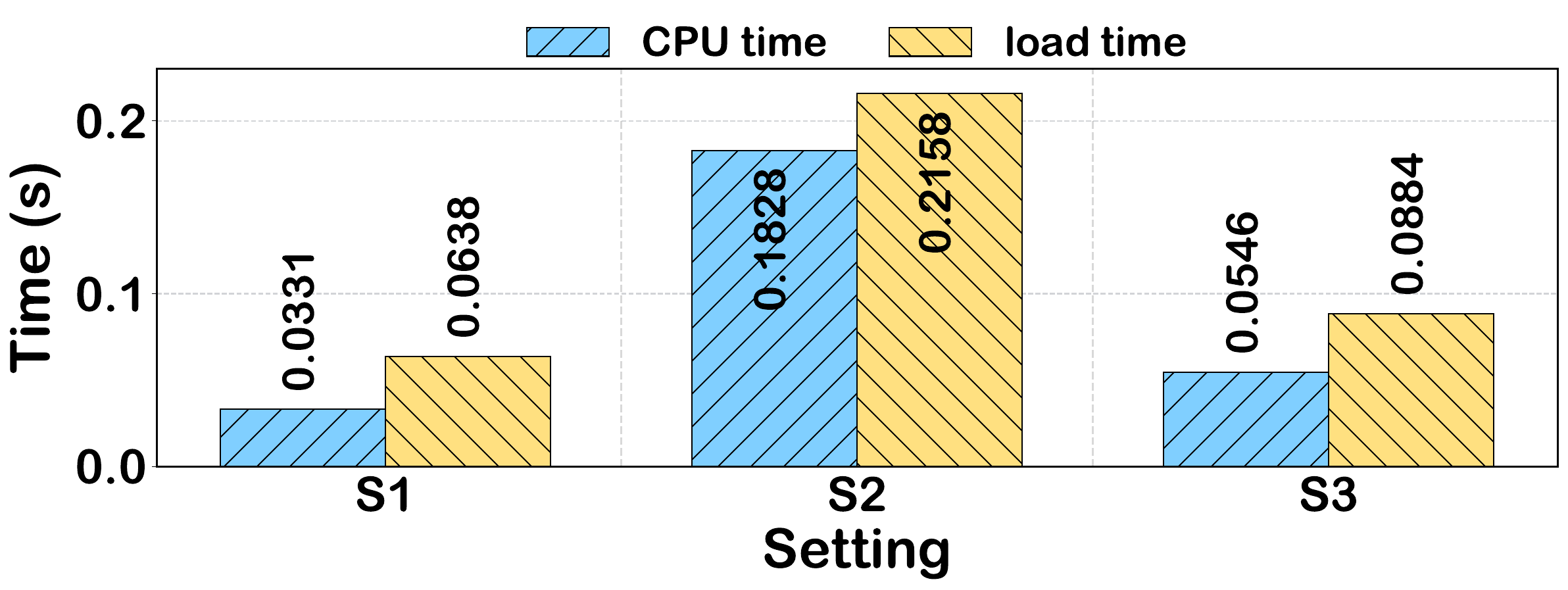}
        \caption{CPU vs PCIe time.}
\label{balancer}

    \end{minipage}
\end{figure*}

\vspace{1mm}
\noindent
\textbf{Prefilling performance.}
Our method consistently achieves a 11\% reduction in TTFT compared to the best existing approach on average, as shown in Fig. \ref{prefillingtime}.  
While our method still outperforms others, this improvement stems from the pipelining strategy across CPU, GPU, and loading processes rather than enhanced expert utilization in GPU memory.  
The prefilling stage differs from the decoding stage: expert activation in decoding is relatively sparse, so improved expert utilization in GPU memory can enhance overall TPOT. 
In contrast, expert activation during prefilling is inherently dense, with nearly all experts in the GPU requiring computation. Thus, increasing expert utilization in GPU memory yields limited gains here.


%

\subsection{Ablation Studies}



Fig. \ref{breakdown} analyzes the role of each component in our method in enhancing overall performance via a stepwise incremental approach (batch size = 1), with each step reducing TPOT.  
The baseline uses LRU expert offloading, but lacks an expert-cache router, prefetching, and our caching strategies.  

\vspace{1mm}
\noindent
\textbf{Cache eviction analysis.}
Compared to the baseline, this reduces TPOT by an average of 8\% across settings and increases GPU cache ratio by 11\% compared to LRU.  
We add a cache-eviction strategy (labeled +CE), which retains recently high-scoring experts during decoding to facilitate replacing those with the lowest score. 
As shown in Fig. \ref{loadratio}, the cache-eviction strategy reduces the probability that active top-score and low-score experts are not in GPU memory, with a more notable improvement in top-score expert hit rates.
This is because top-score experts, which achieve high scores in the current layer, only occasionally have significantly low scores in the recent few tokens; thus, the cache-eviction strategy prioritizes retaining them.
As shown in Fig. \ref{cachecompare}, even ignoring its benefits to the replacement policy, the cache-eviction strategy outperforms the three traditional caching methods.


\vspace{1mm}
\noindent
\textbf{Expert-cache router analysis.}
Relative to +CE, this lowers TPOT by 20\% and improves GPU cache ratio by 60\%.
Building on +CE, we introduce an expert-cache router (labeled +CR) that partitions active experts into top-score and low-score categories, aiming to replace low-score experts with those already in GPU memory.  
As shown in Fig. \ref{loadratio}, the expert-cache router significantly reduces the probability that active low-score experts are not in GPU memory. 
Notably, its optimization effect grows more pronounced as the proportion of low-score experts in active experts increases (\ie as the low ratio rises).

\vspace{1mm}
\noindent
\textbf{Expert prefetching analysis.}
Compared to +CR, this reduces TPOT by 14\% and increases GPU cache ratio by 12\%. 
We incorporate online prefetching (labeled +Pre), which prefetches top-score experts before computation. These first three steps significantly reduce TPOT by improving the GPU expert cache ratio: as shown in Fig. \ref{loadratio}, the number of top-score active experts requiring loading is drastically reduced via prefetching. %

As depicted in Fig. \ref{pciechange}, PCIe load time increases with prefetching, yet TPOT decreases. The increased PCIe load time results from occasional misidentification of low-score experts as top-score ones, leading to unnecessary fetching.
TPOT reduction is achieved by ensuring that prefetching-related PCIe loads overlap with computation, not interfering with current layer expert loading, thus maximizing prefetching benefits. 
Additionally, Fig. \ref{prefetchaccuracy} indicates our prefetching method attains an average accuracy of about 82\% in predicting top-score experts. Even if a prefetched expert is not top-score, it has a 95\% chance of being active, significantly outperforming methods based solely on residual results. Prefetching only these predicted top-score experts allows the router to utilize them directly, boosting efficiency.

\vspace{1mm}
\noindent
\textbf{CPU-assisted Task Load Scheduling.}
Compared to +Pre, this cuts TPOT by 34\% on average but decreases GPU cache ratio by 3\%. 
We integrate the CPU-assisted scheduling (labeled +BA) to balance CPU usage and load. 
The reduction in TPOT is achieved because CPU computation, though slower than GPU computation, overlaps with load time, allowing CPU assistance without additional overhead and maintaining high expert utilization in GPU memory. 

A slight decrease in the GPU cache ratio occurs as experts that would typically be loaded are computed directly on the CPU, which slows GPU cache updates, yet high utilization is maintained. 
As shown in Fig. \ref{balancer}, the scheduling minimizes the difference between CPU and PCIe times, ensuring CPU computation is fully covered by PCIe load time and maximizing CPU computation's supportive role. However, challenges remain: the significant disparity between CPU computation and load times, with CPU time in S3 being only one-third of load time, necessitates many experts for effective balancing. Additionally, the limited number of active experts in MoE per decoding step hinders consistent balancing of PCIe load and CPU computation times.

\section{Related Work}
This section introduces relevant MoE serving approaches.  
For accelerating MoE inference under GPU memory constraints, approaches are categorized into general LLM inference optimization and MoE-specific expert scheduling, with the latter further divided into online and offline MoE serving.

\noindent
\textbf{Online MoE Serving} manages dynamically changing workloads on edge devices, where loading or computing non-resident expert models significantly impacts inference latency. As workloads vary, the active expert set changes, necessitating experts not in GPU memory. These methods adapt via flexible expert scheduling to sustain acceleration effectiveness. MoE-Infinity \cite{xue2024moe} uses a request-level expert activation matrix for coarse offloading decisions, while HybriMoE \cite{zhong2025hybrimoe} optimizes CPU computation, enhancing CPU-load pipelines to minimize the maximum of CPU and loading times.

\vspace{1mm}
\noindent
\textbf{Offline MoE Serving.} Designed for predetermined workloads, these strategies include MoE-lightning's CGOPipe—a CPU-GPU-I/O pipelining schedule with a performance model to maximize resource utilization \cite{moelight}. MoE-Lightning relies heavily on prior knowledge of the workload's average prompt length and generation length to search for its optimal scheduling policy. If this information is unavailable or deviates significantly from the actual workload, it becomes challenging to maintain a consistent micro-batch size across requests. Furthermore, unexpected variations in sequence lengths can lead to excessive CPU memory usage and KV cache transfer overheads. Ultimately, this disrupts the optimized CPU-GPU pipeline, leading to a drastic reduction in overall throughput.
Expert pruning \cite{pruning} reduces memory usage by removing underutilized experts, tailoring resource management to specific workloads.
Knowledge distillation \cite{distillation} produces compact sparse MoE models;
ComPEFT \cite{yadav2023compeft} demonstrates expert compression without accuracy loss, while MC-SMoE \cite{msmoe} further decomposes merged experts into low-rank and sparse alternatives.
All these works address pre-known workloads. However, since we aim to handle unpredictable workloads on edge devices, we only compare with online MoE methods.


\section{Discussion}
\noindent
\textbf{Expert substitution overcomes UMA constraints.}
Our expert substitution strategy remains directly applicable to Unified Memory Architectures (UMA) found in Apple M-series chips. Although UMA eliminates the discrete PCIe link between the CPU and GPU, the total system memory on edge devices is frequently insufficient to house the full parameters of large-scale MoE models. A typical Mac Studio equipped with 32GB or 64GB of memory cannot accommodate models requiring 100GB or more without heavily relying on SSD swapping, which inherently reverts to a PCIe bottleneck. In these memory-constrained scenarios, substituting low-score experts with those already cached in the unified memory pool largely reduces the frequency of high-latency storage accesses. 


\begin{table}[t]
\centering
\caption{Accuracy Gains from Manually Lowered Scores.}
\label{higheraccuracy}

\begin{tabular}{lccc}
\toprule
Dataset & Original & Score-lower & SMoE($\alpha=0.25$) \\
\midrule
Math\_II\_MCQs (\%) & 71.56 & 74.44 & 72.94 \\
History\_MCQs (\%) & 80.84 & 83.97 & 83.90 \\
Biology\_MCQs (\%) & 81.33 & 82.64 & 87.33 \\
race (\%) & 80.0 & 81.3 & 80.4 \\
WiC (\%) & 60.5 & 60.7 & 60.7 \\
\bottomrule
\end{tabular}
\end{table}

\noindent
\textbf{Accuracy Improvement Analysis}. We attribute the accuracy improvement to the suppression of "noisy" activations, which occurs when the gating scores of low-score experts are significantly lower than those of top-score experts.
To provide concrete evidence, we have conducted a validation experiment where we manually decreased the gating scores of these low-score experts without performing any expert substitution. 
As Table \ref{higheraccuracy} show, we observed similar accuracy gains across several datasets, which confirms the regularization effect in some cases.
However, in the other cases low-score experts are close to the top-score experts, such lowering scores will lead to accuracy loss, which explains why methods like expert pruning suffer from accuracy loss.
Since lowering scores alone cannot speed up inference, expert substitution leverages this noise-filtering to increase GPU cache hit rates.

\noindent
\textbf{Design Shifts from kTransformer \cite{ktransformer}.} We did not build SMoE on kTransformer for three reasons: (1) Hardware portability: kTransformer relies on high-end CPU features (\eg AMX and AVX-512) often unavailable on edge devices. (2) Design strategy: kTransformer relies on CPU computation, while SMoE keeps experts GPU-resident with minimal CPU use, removing reliance on high-end CPU. In practice, edge devices rarely have extremely strong CPUs but weak GPUs. (3) Accuracy isolation: We use lossless inference to isolate the accuracy impact of our replacement strategy from quantization noise. SMoE is orthogonal to quantization and can integrate with frameworks like kTransformer.

\noindent
\textbf{Selection of Baselines \& SOTA MoE Systems.} 
These systems address different problems and are orthogonal to substitution. (1) D²MoE \cite{wang2025d2moe} involves mixed quantization and workload-based retraining; in contrast, SMoE is plug-and-play and reduces transfers. (2) ExpertFlow’s \cite{he2024expertflow} routing predictors scale poorly with increasing domains and expert counts. (3) EC2MoE \cite{yang2025ec2moe} optimizes for end-to-cloud bandwidth, unlike our PCIe-based environment.

\section{Conclusion}
In conclusion, we present an approach for deploying MoE LLMs on edge devices by substituting low-importance active experts with functionally similar ones already cached in GPU memory, thereby preserving accuracy. We establish a robust CPU-GPU-load pipeline system named SMoE, providing an effective solution for LLM deployment on edge.

\section{Acknowledgements}
This work was supported in part by the National Key R\&D Program of China under Grant No. 2023YFB4502400, in part by the Fundamental and Interdisciplinary Disciplines Breakthrough Plan of the Ministry of Education of China under Grant No. JYB2025XDXM901, in part by the National Natural Science Foundation of China (NSFC) under Grants No. 62272223, U22A2031, and 62402212, U24B20153, in part by the Natural Science Foundation for Young Scientists of Jiangsu Province under Grant No. BK20241245, in part by the Collaborative Innovation Center of Novel Software Technology and Industrialization, Nanjing University, and in part by the Jiangsu High-level Innovation and Entrepreneurship (Shuangchuang) Program.
\appendix
\section{Artifact Appendix}

\subsection{Abstract}

This artifact provides the source code, scripts, and instructions to reproduce the core latency evaluations of SMoE, an algorithm-system co-design that pushes Mixture-of-Experts (MoE) to the edge via expert substitution. Specifically, this artifact focuses on reproducing the Time Per Output Token (TPOT) and GPU cache hit ratio metrics for the Qwen2-57B-A14B model.

\subsection{Artifact check-list (meta-information)}

{\small
\begin{itemize}
  \item {\bf Algorithm: } Expert Substitution, CPU-assisted task load scheduling.
  \item {\bf Program: } SMoE Evaluation Scripts (Python).
  \item {\bf Model: } Qwen2-57B-A14B-Instruct.
  \item {\bf Dataset: } Gaokao, triviaqa, WiC, Race-mid, gsm8k.
  \item {\bf Hardware: } Single NVIDIA A6000 GPU (48GB) on PCIe Gen 4 with Intel Xeon Gold 6444Y (Setting S3).
  \item {\bf Metrics: } TPOT (Time Per Output Token) and GPU cache hit
  \item {\bf Output: } Logs including TPOT results corresponding to Figure 12 and GPU cache hit corresponding to Figure 13.
  \item {\bf Experiments: } Decoding performance evaluation at batch size = 1.
  \item {\bf How much CPU memory required (approximately)?: } 150GB (The Qwen2-57B-A14B model alone requires 107GB)
  \item {\bf Publicly available?: } Yes (\url{https://github.com/goingshr/SMoE}).
\end{itemize}
}

\subsection{Description}

\subsubsection{How to access}
The source code and evaluation scripts are available at: \url{https://github.com/goingshr/SMoE} or \url{https://doi.org/10.6084/m9.figshare.31982136}.

\subsubsection{Hardware dependencies}
To reproduce the Setting S3 experiments, a single NVIDIA A6000 GPU (48GB) is required. The system should utilize a PCIe 4.0 interface. As specified, the host system requires approximately 150GB of CPU memory to handle the model weights during the loading and offloading processes.

\subsubsection{Software dependencies}
Our implementation relies on Python and several specific packages. A complete list of dependencies and setup instructions is available in the repository's \texttt{README.md}.

\subsubsection{Datasets}
The experiments utilize standard benchmarks: Gaokao, triviaqa, WiC, Race-mid, and gsm8k. Evaluation scripts will automatically download the required subsets.

\subsubsection{Models}
The Qwen2-57B-A14B-Instruct model can be downloaded from HuggingFace.

\subsection{Installation}

The complete source code and environment setup instructions are hosted in our GitHub repository. We delegate the step-by-step installation guide to the repository's \texttt{README.md}. 

\subsection{Experiment workflow}
The experiment workflow is fully automated via provided shell scripts, which trigger inference across the five datasets under specified batch sizes.
\subsection{Evaluation and expected results}

Executing the workflow generates comprehensive execution logs. The primary evaluation metrics are the average Time Per Output Token (TPOT) and the GPU cache hit ratio. 

Our scripts automatically parse these logs to extract both metrics, which are expected to demonstrate significant latency reductions consistent with Figures 12 and 13 (Setting S3) in the paper. For a complete reproducibility guide—including script execution, output parsing, and mapping terminal results to the paper's figures—please refer to the repository's \texttt{README.md}.
\subsection{Experiment customization}
Users can modify the `--alpha` parameter in the execution script to adjust the expert substitution threshold. For the Setting S3 reproduction, $\alpha$ is default to 0.25.
\newpage
\bibliographystyle{IEEEtranS}
\bibliography{sample-base}
\end{document}